\documentclass[10pt,twocolumn,letterpaper]{article}

\usepackage[pagenumbers]{cvpr} % To force page numbers, e.g. for an arXiv version

\usepackage{amsmath}
\usepackage{amssymb}
\usepackage{mathrsfs}
\usepackage{multirow}
\usepackage{pifont}
\usepackage{color}
\usepackage{algorithm}

\usepackage[dvipsnames]{xcolor}

\definecolor{cvprblue}{rgb}{0.21,0.49,0.74}
\usepackage[pagebackref,breaklinks,colorlinks,citecolor=cvprblue]{hyperref}

\def\paperID{9175} % *** Enter the Paper ID here
\def\confName{CVPR}
\def\confYear{2024}

\title{SmartEdit: Exploring Complex Instruction-based Image Editing \\ with Multimodal Large Language Models}

\author{
	Yuzhou Huang$^{*1,2^{\#}}$ \hspace{9pt} Liangbin Xie$^{*2,3,5^{\#}}$ \hspace{9pt} Xintao Wang$^{2,4}$$^\dagger$ \hspace{9pt} Ziyang Yuan$^{2,8^{\#}}$ \hspace{9pt} Xiaodong Cun$^{4}$ \hspace{9pt}\\
    Yixiao Ge$^{2,4}$ \hspace{9pt} Jiantao Zhou$^{3}$ \hspace{9pt} Chao Dong$^{5,7}$ \hspace{9pt} Rui Huang$^{6}$ \hspace{9pt} Ruimao Zhang$^{1}$$^\dagger$ \hspace{9pt} Ying Shan$^{2,4}$ \\
	\vspace{-0.05cm}
\small$^1$School of Data Science, Shenzhen Research Institute of Big Data, The Chinese University of Hong Kong, Shenzhen, China \hspace{5pt} \\ 
\small$^2$\textbf{ARC Lab, Tencent PCG}
\hspace{5pt}\small$^3$University of Macau
 \hspace{5pt} \small$^4$Tencent AI Lab \hspace{5pt}
 \\
\small$^5$Shenzhen Institute of Advanced Technology
\hspace{5pt}\small$^6$School of Science and Engineering, The Chinese University of Hong Kong, Shenzhen, China \hspace{5pt} 
\\
\small$^7$Shanghai Artificial Intelligence Laboratory \hspace{5pt} \small$^8$Tsinghua University\hspace{5pt} \\
\url{https://github.com/TencentARC/SmartEdit}
}

\begin{document}

\twocolumn[{%
\renewcommand\twocolumn[1][]{#1}%
\maketitle
\begin{figure}[H]
\vspace{-1cm}
\hsize=\textwidth
\centering
% \fbox{\rule{0pt}{4in} \rule{0.9\linewidth}{0pt}}
\includegraphics[width=6.9 in]{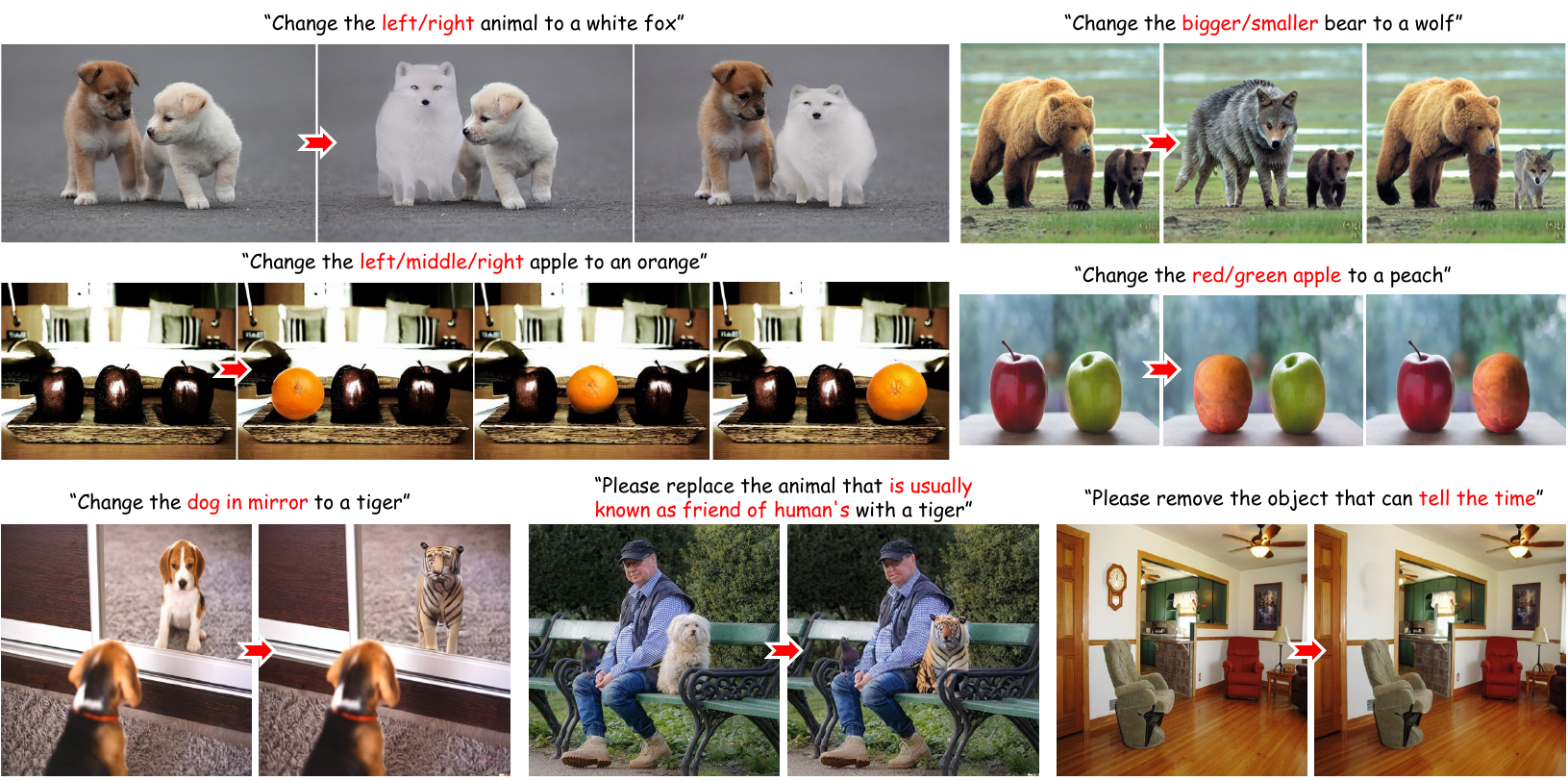}
%\vspace{-0.8cm}
\caption{We propose SmartEdit, an instruction-based image editing model that leverages Multimodal Large Language Models (MLLMs) to enhance the understanding and reasoning capabilities of instruction-based editing methods. With the specialized design, our SmartEdit is capable of handling complex understanding (the instructions that contain various object attributes like location, relative size, color, and in or outside the mirror) and reasoning scenarios.}
\label{fig:teaser} 
\end{figure}
}]

\maketitle
\let\thefootnote\relax\footnotetext{$^*$ Equal contribution\hspace{3pt} \hspace{5pt}$^\dagger$ Corresponding author\hspace{5pt} $^{\#}$ Interns in ARC Lab, Tencent PCG
}

% \begin{document}
% \maketitle
% \let\thefootnote\relax\footnotetext{$^*$ Interns in ARC Lab, Tencent PCG \hspace{3pt} $^\dagger$ Corresponding author.
% }

% \begin{figure*}[t]
% \centering
% \small 
% \begin{minipage}[t]{0.9\linewidth}
% \centering
% \includegraphics[width=1\columnwidth]{figs/teasor.pdf}
% \end{minipage}
% \centering
% % \vspace{2pt}
% \caption{Visual effects of SmartEdit on Reason-Edit dataset (mainly on the complex understanding scenarios). It can be seen that for complex understanding scenarios (e.g., multiple objects, color, relative size, and mirror), SmartEdit has good instruction-based editing effects.}
% % \vspace{-10pt}
% \label{fig:bim} 
% \end{figure*}

\begin{abstract}
Current instruction-based editing methods, such as InstructPix2Pix, often fail to produce satisfactory results in complex scenarios due to their dependence on the simple CLIP text encoder in diffusion models. 
To rectify this, this paper introduces SmartEdit, a novel approach to instruction-based image editing that leverages Multimodal Large Language Models (MLLMs) to enhance their understanding and reasoning capabilities. 
However, direct integration of these elements still faces challenges in situations requiring complex reasoning. 
To mitigate this, we propose a Bidirectional Interaction Module that enables comprehensive bidirectional information interactions between the input image and the MLLM output.
During training, we initially incorporate perception data to boost the perception and understanding capabilities of diffusion models. Subsequently, we demonstrate that a small amount of complex instruction editing data can effectively stimulate SmartEdit's editing capabilities for more complex instructions. 
We further construct a new evaluation dataset, Reason-Edit, specifically tailored for complex instruction-based image editing.
Both quantitative and qualitative results on this evaluation dataset indicate that our SmartEdit surpasses previous methods, paving the way for the practical application of complex instruction-based image editing.
\end{abstract}

\section{Introduction}
\label{sec:intro}

\begin{figure*}[t]
\centering
\small 
\begin{minipage}[t]{0.9\linewidth}
\centering
\includegraphics[width=1\columnwidth]{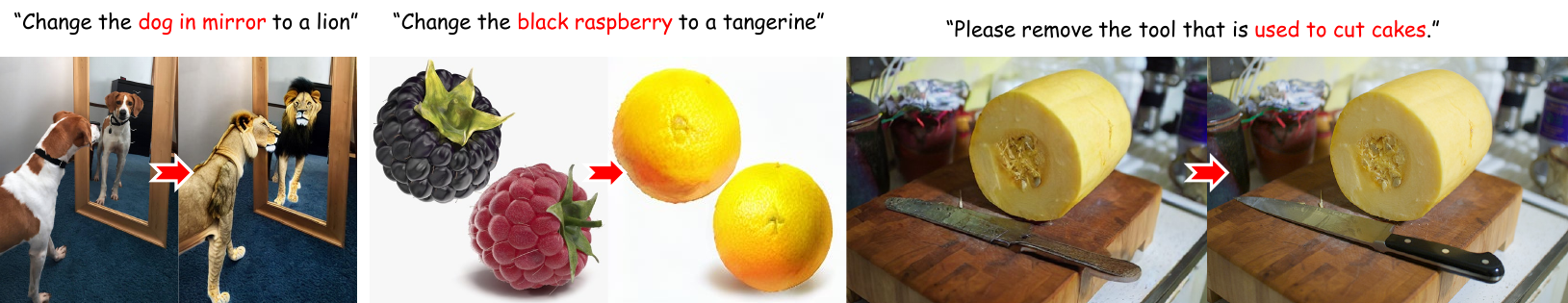}
\end{minipage}
\centering
\vspace{2pt}
\caption{For more complex instructions or scenarios, InstructPix2Pix fails to follow the instructions.}
\vspace{-10pt}
\label{fig:issue} 
\end{figure*}

Text-to-image synthesis~\cite{dhariwal2021diffusion, ho2022classifier, nichol2021glide, ramesh2022hierarchical, saharia2022photorealistic, rombach2022high} has experienced significant advancements in recent years, thanks to the development of diffusion models.
These methods have enabled the generation of images that are not only consistent with natural language descriptions but also align with human perception and preferences, marking a substantial leap forward in the field. 
Instruction-based image editing methods~\cite{brooks2023instructpix2pix, zhang2023magicbrush}, represented by InstructPix2Pix, leverage pre-trained text-to-image diffusion models as priors.
This allows users to conveniently and effortlessly modify images through natural language instructions for ordinary users.

% However, when faced with more complex instructions (e.g., those requiring an understanding of intricate instructions and images (e.g, locations, multiple objects, colors, relative size and et al.), or necessitating complex reasoning (xxx)), previous methods cannot produce satisfying results, and may even fail to follow the instructions, as shown in Fig. xxx. We are intrigued by what impedes the application of these editing methods to such examples. 

While existing instruction-based image editing methods can handle simple instructions effectively, they often fall short when dealing with complex scenarios, which require the model to have a more powerful understanding and reasoning capabilities. As depicted in Fig.~\ref{fig:teaser}, there are two common types of complex scenarios. The first is when the original image contains multiple objects, and the instruction modifies only one of these objects through certain attributes (such as \emph{location, relative size, color, in or outside the mirror}). The other is when world knowledge is needed to identify the object to be edited (such as \emph{the object that can tell the time}). We define these two types as complex understanding scenarios and complex reasoning scenarios, respectively.
%these complex scenarios can mainly be classified into: 
% However, for more complex instructions or scenarios (Fig.~\ref{fig:issue}), 
%(e.g., those requiring an understanding of complex instructions and images, or requiring complex reasoning), 
% existing instruction-based editing methods cannot produce satisfying results, and may even fail to follow the instructions. 
% Unlike simple scenarios or instructions, the scenarios depicted in Fig.~\ref{fig:teaser} require the instruction-based editing model to have more powerful understanding and reasoning capabilities. 
% These scenarios can mainly be classified into 
%1) complex understanding scenarios, such as \emph{multiple objects, color, relative size, and mirror \etal.}; 2) complex reasoning scenarios, such as \emph{object that can tell the time}. 
%
Handling these two scenarios is crucial for practical instruction editing, but existing instruction-based image editing methods probably fail in these scenarios (as shown in Fig.~\ref{fig:issue}). 
In this paper, we attempt to identify the reasons why existing instruction-based image editing methods fail in these scenarios, and try to tackle the challenge in these scenarios.

% We are surprised by these results and are curious about what is hindering the successful application of instruction editing methods in these scenarios.

% In practical instruction editing, common scenarios that require complex reasoning and understanding can mainly be classified into 1) understanding scenarios, such as \emph{multiple objects, color, relative position, and mirror et al.}; 2) complex reasoning, such as \emph{objects blocking the sun}.
% It is crucial for instruction-based editing methods to handle these scenarios well in practical use, yet existing instruction editing methods have not addressed these scenarios. Furthermore, when we apply these instruction-based editing methods to these scenarios, we find that these methods fail in most cases. We are curious about what is hindering the successful application of instruction editing methods in these scenarios. At the same time, we hope to solve this challenging problem.

% In this paper, we conduct comprehensive investigations into this problem and analyze common scenarios that require complex understanding and reasoning. We categorize these scenarios into 1) understanding complex instructions and scenarios, such as locations, multiple objects, colors, and relative size; and 2) complex reasoning, such as xxx. We attempt to address these challenging issues.

The first reason why existing instruction-based image editing methods fail in these scenarios is that they typically rely on a simple CLIP text encoder~\cite{radford2021learning} in diffusion models (\eg, Stable Diffusion) to process the instructions. 
Under this circumstance, these models struggle to 1) understand and reason the instructions, and 2) integrate the image to comprehend the instructions. To address these limitations, we introduce the Multimodal Large Language Model (MLLM) (\eg, LLaVA)~\cite{liu2023visual, zhu2023minigpt} into instruction-based editing models. 
Our method, SmartEdit, jointly optimizes MLLMs and diffusion models, leveraging the powerful reasoning capabilities of MLLMs to facilitate instruction-based image editing task.

While substituting the CLIP encoder in the diffusion model with MLLMs can alleviate some problems, this approach still falls short when it comes to examples that necessitate complex reasoning. 
%This is because the image being edited is incorporated into the UNet of the Stable Diffusion model through a simple input concatenation, and then interacts with the hidden states of the LLM through a classical cross-attention operation. 
This is because the input image to edit (original image) is integrated into the UNet of the Stable Diffusion model through a straightforward concatenation, which is further interacted with MLLM outputs through a cross-attention operation. In this setup, the image feature serves as the query, and MLLM outputs act as the key and value. This means that the MLLM outputs \textit{unilaterally} modulate and interact with the image feature, which affects the results. 
To alleviate this issue, we further propose a Bidirectional Interaction Module (BIM). This module reuses the image information extracted by the LLaVA's visual encoder from the input image. 
It also facilitates a comprehensive bidirectional information interaction between this image and the MLLM output, enabling the model to perform better in complex scenarios.
%The new text information serves as the key and value for the diffusion model, while the new image information supplements the original image and is input into the UNet together with it. The introduction of these two features 

%This module reuses the image feature 
%ViT Encoder in LLaVA to extract information from the edited image and conducts a more comprehensive bidirectional information exchange with the hidden states of the LLM. The image information, after the interaction, is further input into the Stable Diffusion's UNet along with the image to be edited, resulting in improved results.

The second reason contributing to the failure of existing instruction-based editing methods is the absence of specific data.
When solely training on editing datasets, such as the datasets used in Instructpix2pix and MagicBrush, SmartEdit also struggles to handle scenarios requiring complex reasoning and understanding. 
This is because SmartEdit has not been exposed to data from these scenarios. One straightforward approach is to generate a substantial amount of paired data similar to those scenarios.
% Despite the incorporation of the LLM and BIM modules, SmartEdit also struggles to manage scenarios requiring complex reasoning and understanding when solely relying on editing data, such as Instructpix2pix and MagicBrush. 
%one straightforward approach to handle these complex scenarios is to generate a substantial amount of paired data. 
However, this method is excessively expensive because the cost of generating data for these scenarios is high. 
% For instance, MagicBrush requires hiring workers and paying money to create its annotated dataset. Compared to data like MagicBrush, which includes simple instructions and doesn't require much understanding of the image, the cost of producing data for scenarios requiring reasoning and understanding is much higher. Therefore, it's unrealistic to create a paired dataset for complex reasoning and understanding scenarios on the same scale as MagicBrush. 
%We find that current exploration on how to use cost-effective data to further enhance the capabilities of instruction editing models is limited. 
In this paper, we find that there are two keys to enhance SmartEdit's ability to handle complex scenarios. The first is to enhance the perception capabilities of UNet~\cite{ronneberger2015u}, and the second is to stimulate the model capacity in those scenarios with a few high-quality examples. 
Correspondingly, we 1) incorporate the perception-related data (\eg, segmentation) into the model's training.
% which is inspired by InstructDiffusion~\cite{geng2023instructdiffusion}, 
%2) adopt the UNet weights from InstructDiffusion as the initial weights for the UNet in SmartEdit and 3) 
2) synthesize a few high-quality paired data with complex instructions to fine-tune our SmartEdit (similar to LISA~\cite{lai2023lisa}). In this way, SmartEdit not only reduces the reliance on paired data under complex scenarios but also effectively stimulates its ability to handle these scenarios. 

Equipped with the model designs and the data utilization strategy, SmartEdit can understand complex instructions, surpassing the scope that previous instruction editing methods can do. To better evaluate the understanding and reasoning ability of instruction-based image editing methods, we collect the Reason-Edit dataset, which contains a total of $219$ image-text pairs. Note that there is no overlap between the Reason-Edit dataset and the synthesized training data pairs. 
Based on the Reason-Edit dataset, we evaluate existing instruction-based image editing methods comprehensively. Both the quantitative and qualitative results on the Reason-Edit dataset indicate that SmartEdit significantly outperforms previous instruction-based image editing methods.

In summary, our contributions are as follows:

\begin{enumerate}
    \item We analyze and focus on the performance of instruction-based image editing methods in more complex instructions. These complex scenarios have often been overlooked and less explored in past research. %However, understanding and efficiently handling these scenarios is crucial for enhancing the practical usability of instruction-based image editing methods.
    \item We leverage MLLMs to better comprehend instructions. To further improve the performance, we propose a Bidirectional Interaction Module to facilitate the interaction of information between text and image features. 
    \item We propose a new dataset utilization strategy to enhance the performance of SmartEdit in complex scenarios. In addition to using conventional editing data, we introduce perception-related data to strengthen the perceptual ability of UNet in the diffusion process. Besides, we also add a small amount of synthetic editing data to further stimulate the model's reasoning ability.
    % Such a strategy has not been explored currently, but it is very effective in complex instruction-based image editing tasks.}
    \item An evaluation dataset, Reason-Edit, is specifically collected for evaluating the performance of instruction-based image editing tasks in complex scenarios. Both qualitative and quantitative results on Reason-Edit demonstrate the superiority of SmartEdit.
\end{enumerate}

\section{Related Work}
\label{sec:related_work}

\subsection{Image Editing with Diffusion Models.}
Pretrained text-to-image diffusion models~\cite{dhariwal2021diffusion, ho2022classifier, nichol2021glide, ramesh2022hierarchical, saharia2022photorealistic, rombach2022high} can strongly assist image editing task. Instruction-based image editing task~\cite{zhang2023magicbrush, brooks2023instructpix2pix, cao2023masactrl, kawar2023imagic, tumanyan2023plug, hertz2022prompt, geng2023instructdiffusion, ju2023direct} requires users to provide an instruction, which converts the original image to a newly designed image that matches the given instruction.
Some methods can achieve this by utilizing a tuning-free approach.
For example, Prompt-to-Prompt~\cite{hertz2022prompt} suggests modifying the cross-attention maps by comparing the original input caption with the revised caption. MasaCtrl~\cite{cao2023masactrl} converts existing self-attention in diffusion models into mutual self-attention, which can help query correlated local contents and textures from source images for consistency.
%concentrates on executing non-rigid consistent image synthesis and editing, by employing a mask to reduce the ambiguity between foreground and background elements. 
In addition, due to the scarcity of paired image-instruction editing datasets, the pioneering work InstructPix2Pix~\cite{brooks2023instructpix2pix} introduces a large-scale vision-language image editing datasets created by fine-tuned GPT-3~\cite{brown2020language} and Prompt-to-Prompt with stable diffusion, and further fine-tunes the UNet~\cite{ronneberger2015u}, which can edit images by providing a simple instruction. To enhance the editing effect of InstructPix2Pix on real images, MagicBrush~\cite{zhang2023magicbrush} further provides a large-scale and manually annotated dataset for instruction-guided real image editing. 

The recent work, InstructDiffusion~\cite{geng2023instructdiffusion}, also adopts the network design of InstructPix2Pix and focuses on unifying vision tasks in a joint training manner. By taking advantage of multiple different datasets, it can handle a variety of vision tasks, including understanding tasks (such as segmentation and keypoint detection) and generative tasks (such as editing and enhancement). Compared with InstructDiffusion, our primary focus is on the field of instruction-based image editing, especially for complex understanding and reasoning scenarios. In these scenarios, InstructDiffusion typically generates inferior results.

\subsection{LLM with Diffusion Models}
The exceptional open-sourced LLaMA~\cite{touvron2023llama, vicuna2023} significantly enhances the performance of vision tasks with the aid of Large Language Models (LLMs). Pioneering works such as LLaVA and MiniGPT-4~\cite{liu2023visual, zhu2023minigpt} have improved image-text alignment through instruction-tuning. 
While numerous MLLM-based studies have demonstrated their robust capabilities across a variety of tasks, primarily those reliant on text generation (e.g., human-robot interaction, complex reasoning, science question answering, etc.), GILL~\cite{koh2023generating} serves as a bridge between MLLMs and diffusion models. It learns to process images with LLMs and is capable of generating coherent images based on the input texts. SEED~\cite{ge2023planting} presents an innovative image tokenizer to enable LLM to process and generate images and text concurrently. SEED-2~\cite{ge2023making} further refines the tokenizer by aligning the generation embedding with the image embedding of unCLIP-SD, which allows for better preservation of rich visual semantics and reconstruction of more realistic images. Emu~\cite{sun2023generative} can be characterized as a multimodal generalist, trained with the next-token-prediction objective. 
% This training not only equips it to perform image-to-text tasks but also facilitates image-to-image translation and text-to-image tasks, contingent on the provided source image and text prompt, respectively.
CM3Leon~\cite{yu2023scaling} proposes a multi-modal language model that is capable of executing text-to-image and image-to-text generation. It employs the CM3 multi-modal architecture that is fine-tuned on diverse instruction-style data, and utilizes a training method adapted from text-only language models. 
% Inspired by GILL, we propose to further explore complex instruction-based image editing with MLLMs.

% 原来
% While Multimodal Large Language Models (MLLMs) initially showcased their potent capabilities across a range of discriminative tasks, GILL~\cite{koh2023generating} bridges the gap between MLLMs and diffusion models by learning to process images with LLMs.       
% Emu~\cite{sun2023generative} can be considered a multimodal generalist, trained with the next-token-prediction objective. This not only enables it to perform image-to-text tasks, but also facilitates image-to-image translation and text-to-image tasks based on the provided source image and text prompt, respectively. 

\section{Preliminary}
\label{sec:preliminary}

The goal of instruction-based image editing is to make specific modifications to an input image $x$ based on instructions $c_{T}$, resulting in the target image $y$. InstructPix2Pix, which is based on latent diffusion, is a seminal work in this field. For the target image $y$ and an encoder $\mathcal{E}$, the diffusion process introduces noise to the encoded latent $z=\mathcal{E}(y)$, resulting in a noisy latent $z_{t}$, with the noise level increasing over timesteps $t \in T$. A UNet $\epsilon_\delta $ is then trained to predict the noise added to the noisy latent $z_{t}$, given the image condition $c_{x}$ and text instruction condition $c_{T}$, where $c_{x}=\mathcal{E}(x)$. The image condition is incorporated by directly concatenating $c_{x}$ and $z_{t}$. The specific objective of latent diffusion is as follows:

{
\begin{align}
L_{\mathrm{diffusion}}=\mathbb{E}_{\mathcal{E}(y), \mathcal{E}(x), c_{T}, \epsilon \sim \mathcal{N}(0,1), t}[\| \epsilon \nonumber \\ -\epsilon_\delta(t, \mathrm{concat}[z_t, \mathcal{E}(x)], c_T)) \|_2^2
% \tag{1}
\label{eq:diffusion}
\end{align}
}where $\epsilon$ is the unscaled noise, $t$ is the sampling step, $z_{t}$ is latent noise at step $t$, $\mathcal{E}(x)$ is the image condition, and $c_{T}$ is the text instruction condition. The $\mathrm{concat}$ corresponds to the concatenation operation.

Although InstructPix2Pix has some effectiveness in instruction editing, its performance is limited when dealing with complex understanding and reasoning scenarios. To address this issue, we introduce a Multimodal Large Language Model (MLLM) into the network architecture and propose a Bidirectional Interaction Module (BIM) to implement bidirectional information interaction between the MLLM output and image information. In addition, we also explore the data utilization strategy and find that perception-related data and a small amount of complex editing data are crucial for enhancing model's performance. We provide detailed descriptions of these aspects in the next section.

\begin{figure*}[t]
\centering
\small 
\begin{minipage}[t]{\linewidth}
\centering
\includegraphics[width=1.0\columnwidth]{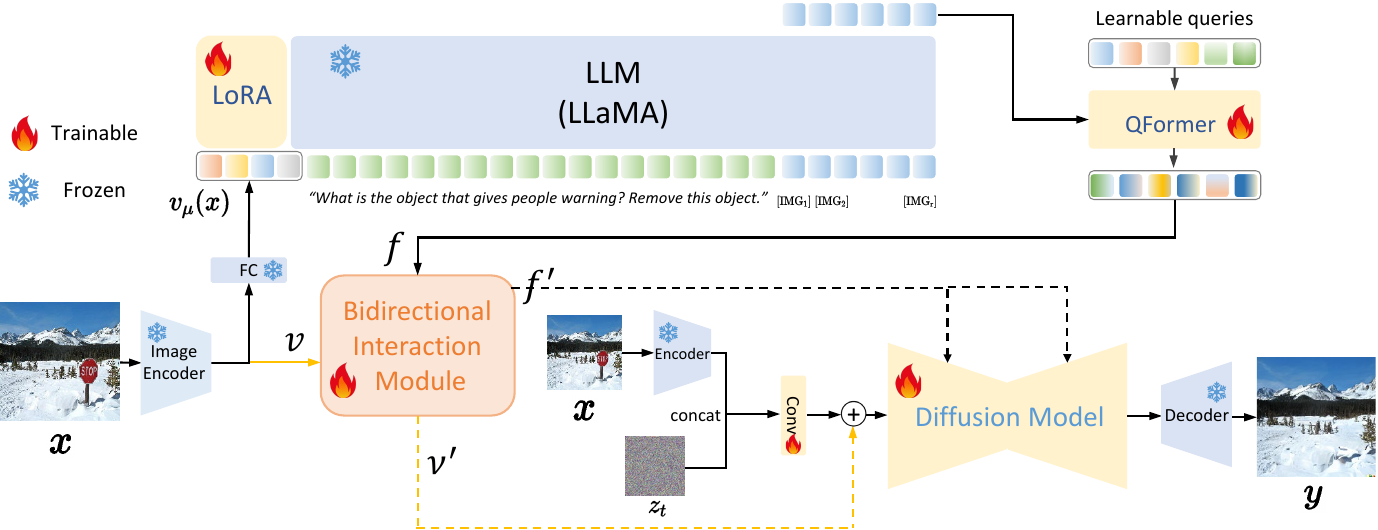}
\end{minipage}
\centering
\vspace{-8pt}
\caption{The overall framework of SmartEdit. For the instruction, we first append the $r$ $[\mathrm{IMG}]$ tokens to the end of instruction $c$. Together with image $x$, they will be sent into LLaVA, which can then obtain the hidden states corresponding to these $r$ $[\mathrm{IMG}]$ tokens. Then the hidden state is sent into the QFormer and gets feature $f$. Subsequently, the image feature $v$ output by the image encoder $E_{\phi}$ interacts with $f$ through a bidirectional interaction module (BIM), resulting in $f'$ and $v'$. The $f'$ and $v'$ are input into the diffusion models to achieve the instruction-based image editing task.
}
% \vspace{-10pt}
\label{fig:framework} 
\end{figure*}
\section{Method}
\label{sec:method}

In this paper, we introduce SmartEdit, specifically designed to handle complex instruction editing scenarios. In this section, we first provide a detailed overview of the framework of SmartEdit (Section~\ref{method:framework}). Then, we delve into the Bidirectional Interaction Module (Section~\ref{method:bim}). In Section~\ref{method:traindata}, we discuss how to enhance the perception and understanding capabilities of UNet in the diffusion model and stimulate the ability of Multimodal Large Language Model (MLLM) to handle complex scenarios. Finally, We introduce Reason-Edit, which is primarily used to evaluate the ability of instruction-based image editing methods toward complex scenarios. (Section~\ref{method:benchmark}). 
% Based on this, we have built a benchmark for existing methods (Section~\ref{method:benchmark}).

\subsection{The Framework of SmartEdit}
\label{method:framework}

Given an image $x$ and instruction $c$, which is tokenized as ($s_{1}$, ..., $s_{T}$), our goal is to obtain the target image $y$ based on $c$. As shown in Fig~\ref{fig:framework}, the image $x$ is first processed by the image encoder and FC layer, resulting in $v_{\mu}(x)$. Then $v_{\mu}(x)$ is sent into the LLM along with the token embedding ($s_{1}$, ..., $s_{T}$). The output of the LLM is discrete tokens, which cannot be used as the input for subsequent modules. Therefore, we take the hidden states corresponding to these discrete tokens as the input for the following modules.  To jointly optimize LLaVA and the diffusion model, following GILL~\cite{koh2023generating}, we expand the original LLM vocabulary with $r$ new tokens $[\mathrm{IMG}_{1}]$,...,$[\mathrm{IMG}_{r}]$ and append the $r$ $[\mathrm{IMG}]$ tokens to the end of instruction $c$. To be specific, we incorporate a trainable matrix $\mathbf{E}$ into the embedding matrix of the LLM, which represents the $r$ $[\mathrm{IMG}]$ token embeddings. Subsequently, we minimize the negative log-likelihood of generated $r$ $[\mathrm{IMG}]$ tokens, conditioned on tokens that have been generated previously:

% Following GILL [], 

% we expand the original LLM vocabulary with $r$ new tokens $[\mathrm{IMG}_{1}]$,...,$[\mathrm{IMG}_{r}]$. The hidden states $h$ that the LLM produces for these tokens will replace the features produced by the original clip text encoder in the diffusion model. 

% Specifically, we incorporate a trainable matrix $\mathbf{E}$ into the embedding matrix of the LLM, which represents the $r$ $[\mathrm{IMG}]$ token embeddings. To efficiently transfer the information from the image $x$ and the instruction $c$ to the hidden states associated with the $r$ $[\mathrm{IMG}]$ tokens, we append the $r$ $[\mathrm{IMG}]$ tokens to the end of instruction $c$ during the training process. Subsequently, we minimize the negative log-likelihood of generating $r$ $[\mathrm{IMG}]$ tokens, conditioned on tokens that have been generated previously:

{\small
\begin{align}
L_{\mathrm{LLM}}(c)=-\sum_{i=1}^r \log p_{\left\{\theta \cup \mathbf{E}\right\}} ([\operatorname{IMG}_{i}] \mid v_{\mu}(x),\nonumber \\ s_{1}, ..., s_{T}, [\operatorname{IMG}_{1}], \ldots,[\operatorname{IMG}_{i-1}])
\end{align}
}

The majority of parameters $\theta$ in the LLM are kept frozen and we utilize LoRA~\cite{hu2021lora} to carry out efficient fine-tuning. We take the hidden states $h$ corresponding to the $r$ $[\mathrm{IMG}]$ tokens as the input for the next module.

% Since the text space corresponding to the diffusion model is the clip text encoder space, we need to align the hidden states $h$ of the LLM with the clip text encoder space. 

Considering the discrepancy between the feature spaces of the hidden states in the LLM and the clip text encoder, we need to align the hidden states $h$ to the clip text encoder space. Inspired by BLIP2~\cite{li2023blip} and DETR~\cite{carion2020end}, we adopt a QFormer $Q_{\beta}$ with $6-$layers of transformer~\cite{vaswani2017attention} and $n$ learnable queries, obtaining feature $f$. Subsequently, the image feature $v$ output by the image encoder $E_{\phi}$ interacts with $f$ through a bidirectional interaction module (BIM), resulting in $f'$ and $v'$. The process mentioned above is represented as:

\begin{equation}
\begin{aligned}
& h=\mathrm{LLaVA}(x, c), \\
& f=Q_\beta\left(h\right), \\
& v=E_\phi(x), \\
& f', v' = \mathrm{BIM}(f, v)
\end{aligned}
\end{equation}

For the diffusion model, 
%regarding the image condition $x$, we adopt the original design of Instructpix2pix, 
following the design of Instructpix2pix, we concat the encoded image latent $\mathcal{E}(x)$ and noisy latent $z_{t}$. Unlike Instructpix2pix, we use $f'$ as the key and value in UNet, and combine $v'$ into the features before entering UNet in a residual manner.The specific process can be formulated as:

{
\begin{align}
L_{\mathrm{diffusion}}=\mathbb{E}_{\mathcal{E}(y), \mathcal{E}(x), c_{T}, \epsilon \sim \mathcal {N}(0,1), t}[\| \epsilon \nonumber \\ -\epsilon_\delta(t, \mathrm{concat}[z_t, \mathcal{E}(x)] + v', f')) \|_2^2
\end{align}
}
To keep consistency with equation~\ref{eq:diffusion}, we omit the Conv operation here.

% and the UNet is trainable during training.

\begin{figure}[t]
\centering
\small 
\begin{minipage}[t]{0.8\linewidth}
\centering
\includegraphics[width=1\columnwidth]{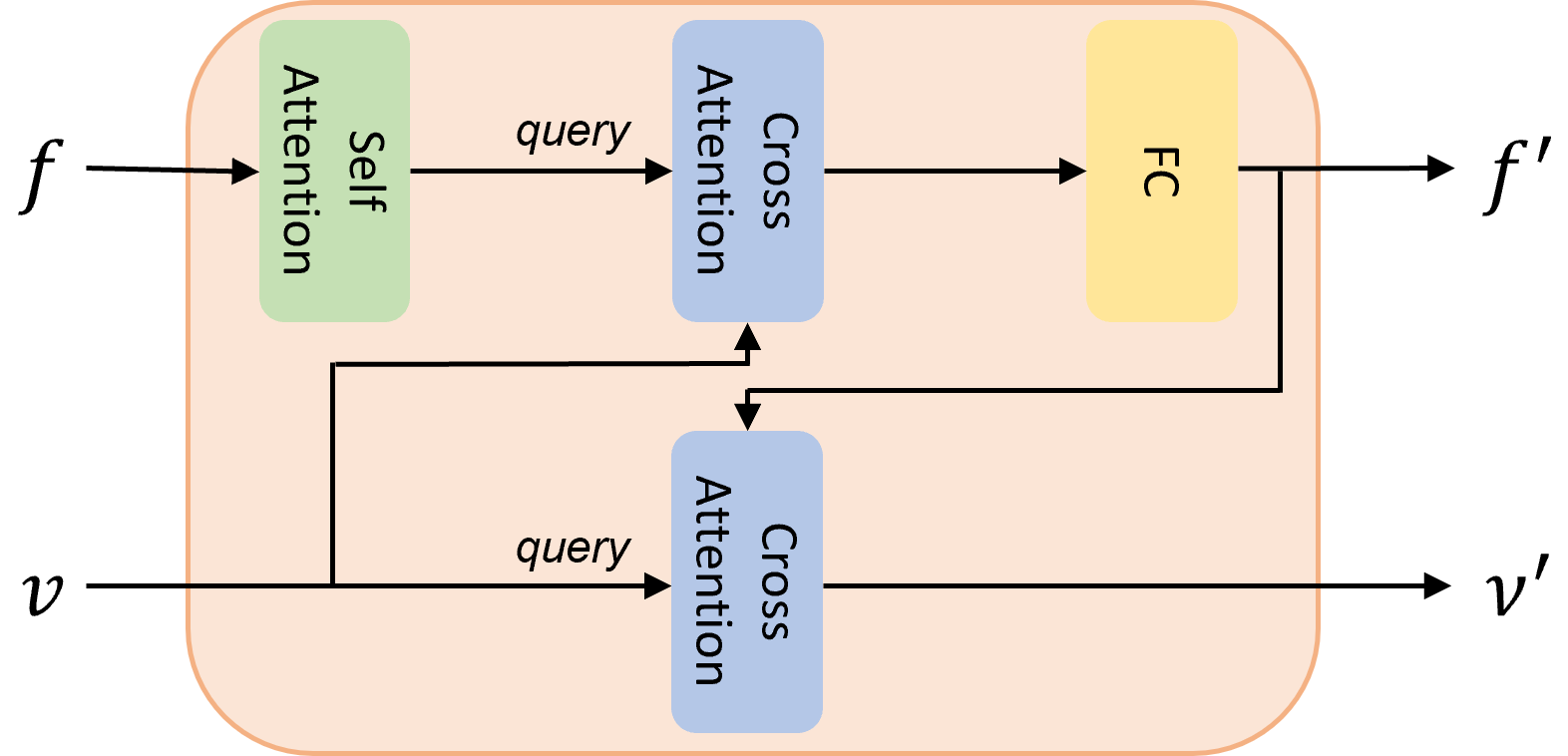}
\end{minipage}
\centering
% \vspace{2pt}
\caption{The network design of the BIM Module. In this module, the input information $f$ and $v$ will undergo bidirectional information interaction through different cross-attention.}
\vspace{-18pt}
\label{fig:bim_network} 
\end{figure}

\subsection{Bidirectional Interaction Module}
\label{method:bim}

The design of BIM is depicted in Fig~\ref{fig:bim_network}. It includes a self-attention block, two cross-attention blocks, and an MLP layer. The two inputs of BIM are the output $f$ from QFormer and the output $v$ from the image encoder. After bidirectional information interaction between $f$ and $v$, BIM will eventually output $f'$ and $v'$. In BIM, the process begins with $f$ undergoing a self-attention mechanism. After this, $f$ serves as a query to interact with the input $v$, which acts as both key and value, through a cross-attention block. This interaction results in the generation of $f'$ via a point-wise MLP. Following the creation of $f'$, it then serves as both key and value to interact with $v$, which now acts as a query. This second cross-attention interaction leads to the production of $v'$.

% Specifically, in BIM, $f$ first goes through a self-attention process, then serves as a query and interacts with the input $v$ (serving as key and value) through cross-attention, and finally goes through a point-wise MLP to get $f'$. After generating $f'$, $v$ serves as a query and interacts with $f'$ (serving as key and value) through cross-attention, resulting in $v'$.

As discussed in the introduction, the proposed BIM module reuses the image feature and inputs it as supplementary information into UNet. The implementation of two cross-attention blocks in this module facilitates a robust bidirectional information interaction between the image feature and the text feature.
 %there is a sufficient bidirectional information interaction between the image feature and the text feature. 
 Compared to not adopting the BIM module or only fusing the image feature and text feature in one direction, SmartEdit which is equipped with the BIM module yields better results. The experimental comparison of different designs is shown in Section~\ref{sec:ablation_bim}.

% As discussed in the introduction, the proposed BIM module enables extensive information interaction between text features and image features by using them as queries and fusing them with other features. Compared to not fusing information or only fusing information in one direction, this bidirectional information fusion method of BIM yields better results. In addition, considering that the image feature $v'$ has already been fused with the text feature $f'$ and has a good correspondence, we input this image feature as supplementary information into the UNet~\cite{ronneberger2015u}. This can achieve more precise instruction editing. The experimental comparison of different designs is shown in Section~\ref{sec:ablation_bim}.

%In addition, considering that the image feature $v'$ has already been fused with the text feature $h'$ and has a good correspondence, we input this image feature as supplementary information into the UNet. This operation can achieve more precise instruction editing.

\subsection{Dataset Utilization Strategy}
\label{method:traindata}

During the training process of SmartEdit, two primary challenges emerge when solely utilizing datasets gathered from InstructPix2Pix and MagicBrush as the training set. 
The first challenge is that SmartEdit has a poor perception of position and concept. The second challenge is that, despite being equipped with MLLM, SmartEdit still has limited capability in scenarios that require reasoning. In summary, the effectiveness of SmartEdit in handling complex scenarios is limited if it is only trained on conventional editing datasets.
% by , SmartEdit is unable to handle scenarios that require complex understanding and reasoning.
% The first challenge is that the UNet in the diffusion model lacks an understanding of perception and concepts, which hampers effective instruction editing in scenarios that require complex understanding. The second challenge is that SmartEdit has not been exposed to editing data that necessitates reasoning capabilities, which limits its performance in reasoning scenarios, even with the introduction of LLM.
% To address these challenges, two key modifications are made. 
After analyses, 
%carrying out a multitude of experiments and analyses, 
we have identified the causes of these issues. The first issue stems from the UNet in the diffusion model which lacks an understanding of perception and concepts, leading to SmartEdit's poor perception of position and concept. The second issue is that SmartEdit has limited exposure to editing data that requires reasoning abilities, which in turn limits its reasoning capabilities.
%The second issue is tied to SmartEdit's limited exposure to editing data that requires reasoning abilities, resulting in its limited capacity for reasoning despite being equipped with LLM. 
% We further discovered that by making two improvements to the training data, we can effectively solve these two problems. For the first issue, we 

To tackle the first issue, we incorporate the segmentation data into the training set. Such modifications significantly enhanced the perception capabilities of the SmartEdit model.
% emulate InstructDiffusion by incorporating segmentation data into the training set. Furthermore, we use the UNet weights from InstructDiffusion as the initial weights for the UNet in SmartEdit. 
% leverage the weights of InstructDiffusion [], which have been pre-trained on numerous perception datasets such as segmentation datasets. By adopting the UNet weights of InstructDiffusion as the initial weights, the model's understanding of perception and concepts is enhanced. 
Regarding the second issue, we take inspiration from LISA~\cite{lai2023lisa} that a minimal amount of reasoning segmentation data can efficiently activate MLLM's reasoning capacity.
%a project that effectively merges LLM and SAM~\cite{kirillov2023segment} for reasoning segmentation tasks. LISA demonstrates that training a model with LLM, using a minimal amount of segmentation data that necessitates reasoning, can efficiently activate LLM's reasoning capacity. 
Guided by this insight, we establish a data production pipeline and synthesize approximately $476$ paired data (each sample contains an original image, instruction, and the synthetic target image) as a supplement to the training data. This synthetic editing dataset includes two major types of scenarios: complex understanding scenarios and reasoning scenarios. For complex understanding scenarios, the original image contains multiple objects and the corresponding instruction modifies the specific object based on various attributes (i.e., location, color, relative size, and in or outside the mirror). We specifically consider the mirror attribute because it is a typical example that requires a strong understanding of the scene (both inside and outside the mirror) to perform well. For reasoning scenarios, we involve complex reasoning cases that need world knowledge to identify the specific object. The effectiveness of this synthetic editing dataset and the impact of different datasets on the model's performance are detailed in Section~\ref{sec:ablation_data}. The details of the data production pipeline and some visual examples are described in the supplementary material.

% \noindent \textbf{Discussion:} We acknowledge that "Replace" and "Remove" cannot cover the entire range of instruction editing. However, considering that it's relatively convenient to produce synthetic data pairs under these two types of instructions, and many instruction edits can be achieved through a combination of "Replace" and "Remove", the majority of instructions in our synthesized data set are of these two types. In the future, we plan to add more instructions to our synthetic paired data set.

% Specifically, a small amount of corresponding paired data is created for each category, such as position, multiple objects, color, relative position, mirror, and categories requiring reasoning. The specific number for each category is (). Some examples of the data produced, primarily using the instructions 'Remove' and 'Replace', are included.
% The effectiveness of these modifications and the impact of different datasets on the model's performance will be detailed in Section 5.3. The details of the data production pipeline will be described in the supplementary material.

\subsection{Reason-Edit for Better Evaluation}
\label{method:benchmark}

To better evaluate existing instruction editing methods and SmartEdit's capabilities in complex understanding and reasoning scenarios, we collect an evaluation dataset, Reason-Edit. Reason-Edit consists of 219 image-text pairs. Consistent with the synthetic training data pairs, 
%which are designed to stimulate LLM in image editing, 
Reason-Edit is also categorized in the same manner. Note that there is no overlap between the data in Reason-Edit and the training set. With Reason-Edit, we can thoroughly test the performance of instruction-based image editing models in terms of understanding and reasoning scenarios. We hope more researchers will pay attention to the capabilities of instruction-based image editing models from these perspectives, thereby fostering the practical application of instruction-based image editing methods. 

% Current instruction-based image editing methods do not focus on complex understanding and reasoning scenarios, and there is also a lack of corresponding datasets to evaluate the capabilities of instruction-based image editing models. To address this, we collected xxx images from the internet to construct an evaluation set. Based on this evaluation set, we evaluated existing methods and established corresponding benchmark results. Our assessment of model capabilities is divided into several aspects. The small amount of complex instruction editing data and evaluation set we created will be open-sourced. We hope that more researchers will pay attention to the capabilities of instruction-based image editing models in these areas, thereby promoting the development of this field.
% \vspace{-2mm}
\section{Experiments}
\label{sec:experiments}
% \vspace{-2mm}
\subsection{Experimental Setting}

\noindent \textbf{Training Process.}
% \textcolor{red}{xxxxx}
The training process of SmartEdit is divided into two main stages. In the first stage, the MLLM is aligned with the CLIP text encoder~\cite{radford2021learning} using the QFormer~\cite{li2023blip}. 
%The large corpus CC12M~\cite{changpinyo2021conceptual} is used as a medium for language alignment, which helps in aligning the feature spaces of the LLM and CLIP text encoder. 
In the second stage, we optimize SmartEdit. To be specific, the weights of LLaVA are frozen and LoRA~\cite{hu2021lora} is added for efficient fine-tuning. Since InstructDiffusion also trains on the segmentation dataset, for convenience, we directly use its weights as the initial weights for the diffusion model in SmartEdit. During the second stage, QFormer, BIM module, LoRA, and UNet~\cite{ronneberger2015u} in the diffusion model are fully optimized. 

\noindent \textbf{Network Architecture.} For the Large Language Model with visual input (e.g., LLaVA), we choose LLaVA-1.1-7b and LLaVA-1.1-13b as the base model. During training, the weights of LLaVA are frozen and we add LoRA for efficient fine-tuning. In LoRA, the values of the two parameters, dim and alpha, are 16 and 27, respectively. We expand the original LLM vocabulary with $32$ new tokens. The QFormer is composed of $6$ transformer~\cite{vaswani2017attention} layers and $77$ learnable query tokens. In the BIM module, there is a self-attention block, two cross-attention blocks, and a Multilayer Perceptron (MLP) layer.

\noindent \textbf{Implementation Details.} During the first stage of training, the AdamW optimizer~\cite{loshchilov2017decoupled} is used, and the learning rate and weight decay parameters are set to 2e-4 and 0, respectively. 
% Additionally, the value of warp a warm-up ratio of 0.04 is adopted. 
The training objectives at this stage are the combination of the mse loss between the output of LLaVA and clip text encoder, and the language model loss. The weights of both losses are 1. In the second stage, we also adopt the AdamW optimizer. The values of learning rate, weight decay, and warm-up ratio were set to 1e-5, 0, and 0.001, respectively. In this phase, the loss function is composed of two parts: the language model loss and the diffusion loss. The ratio of these two losses is 1:1.

% All our LLaVA-1.1-7b experiments are conducted on 8 NVIDIA-40G A100 GPUs and LLaVA-1.1-13b experiments are conducted on 8 NVIDIA-80G H800 GPUs. The training is facilitated using the Hugging Face Trainer with the deepspeed~\cite{rasley2020deepspeed} engine and ZeRO-2 configuration. The AdamW~\cite{loshchilov2017decoupled} optimizer is adopted, with the learning rate and weight decay set to 1e-5 and 0, respectively. We also use WarmupLR as the learning rate scheduler, with a warmup ratio set to 0.001. For the training processes, the weights of the LLM Loss and the Diffusion Loss are both set to 1.0. The batch size for each device is set to 4, with a gradient accumulation step of 4.

\noindent \textbf{Training Datasets.} In the first stage, we utilize the extensive corpus CC12M~\cite{changpinyo2021conceptual} as our primary data source. In the second stage, the training data can be divided into $4$ categories: (1) segmentation datasets, which include COCOStuff~\cite{caesar2018coco}, RefCOCO~\cite{yu2016modeling}, GRefCOCO~\cite{liu2023gres}, and the reasoning segmentation dataset from LISA~\cite{lai2023lisa}; (2) editing datasets, which involve InstructPix2Pix and MagicBrush; (3) visual question answering (VQA) dataset, which is the LLaVA-Instruct-150k dataset~\cite{liu2023visual}; (4) synthetic editing dataset, where we collect a total of 476 paired data for complex understanding and reasoning scenarios.

\noindent \textbf{Evaluation Metrics.} 
% To better evaluate the effects of the edited images, we divide the edited images into foreground and background based on the area of the edited object. Our aim during the editing process is to alter only the foreground while keeping the background intact. For the background area, we use three metrics: PSNR, SSIM, and LPIPS. These metrics help us measure how well we have preserved the background during the editing process. 
As we hope to only change the foreground of the image while keeping the background unchanged during the editing process, we adopt three metrics for the background area: PSNR, SSIM, and LPIPS~\cite{hore2010image, zhang2018unreasonable}. For the foreground area, we calculate the CLIP Score~\cite{radford2021learning} between the foreground area of the edited image and the GT label. The GT label is annotated manually. Among these four metrics, except for LPIPS where lower is better, the other three metrics are higher the better. While these metrics can reflect the performance to a certain extent, they are not entirely accurate. To provide a more accurate evaluation of the effects of edited images, we propose a metric for assessing editing accuracy. Specifically, we hire four workers to manually evaluate the results of these different methods on Reason-Edit. The evaluation criterion is whether the edited image aligns with the instruction. After obtaining the evaluation results from each worker, we average all the results to get the final metric result, which is Instruction-Alignment (Ins-align).

\begin{figure*}[t]
\centering
\small 
\begin{minipage}[t]{0.9\linewidth}
\centering
% \vspace{-20pt}
\includegraphics[width=1\columnwidth]{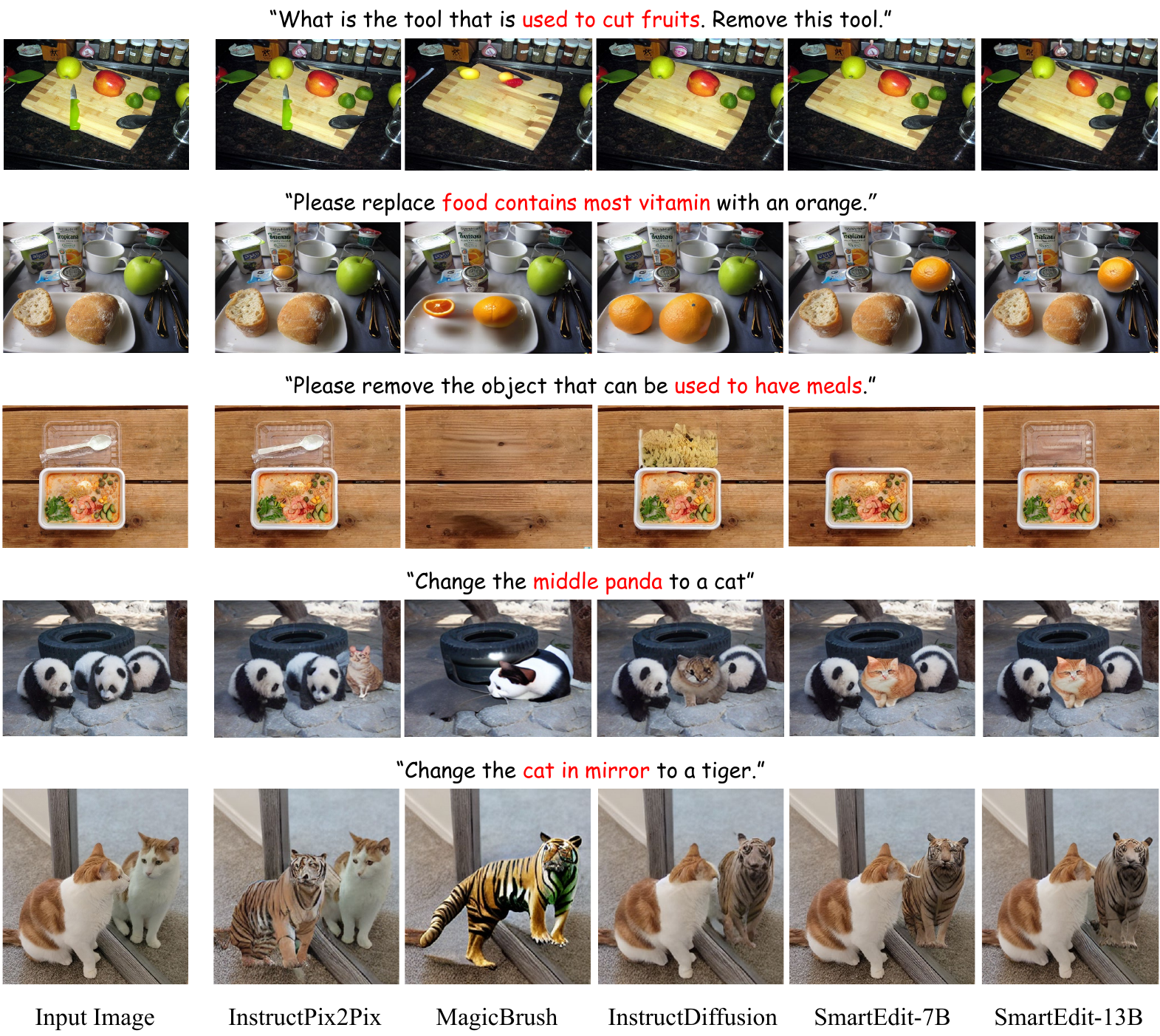}
\end{minipage}
\centering
% \vspace{5pt}
\caption{Qualitative comparison on Reason-Edit. When compared to several existing instruction-based image editing methods that have undergone further fine-tuning on our synthetic editing dataset, our approach demonstrates superior editing capabilities in complex scenarios.}
\label{fig:comparison} 
\end{figure*}

% Please add the following required packages to your document preamble:
% \usepackage{multirow}
\begin{table*}
\centering
\resizebox{1.0\textwidth}{!}{
\begin{tabular}{c|ccccc|ccccc}
\hline
\multirow{2}{*}{Methods} & \multicolumn{5}{c|}{Understanding Scenarios}               & \multicolumn{5}{c}{Reasoning Scenarios}                    \\ \cline{2-11} 
                         & PSNR(dB)$\uparrow$ & SSIM$\uparrow$  & LPIPS$\downarrow$ & CLIP Score$\uparrow$ & \textbf{Ins-align$\uparrow$} & PSNR(dB) & SSIM  & LPIPS & CLIP Score & \textbf{Ins-align$\uparrow$} \\ \hline
InstructPix2Pix          & 21.576   & 0.721 & 0.089 & 22.762     & 0.537              & 24.234   & 0.707 & 0.083 & 19.413     & 0.344              \\
MagicBrush               & 18.120   & 0.68  & 0.143 & 22.620     & 0.290              & 22.101   & 0.694 & 0.113 & 19.755     & 0.283              \\
InstructDiffusion        & 23.258   & 0.743 & 0.067 & 23.080     & 0.697              & 21.453   & 0.666 & 0.117 & 19.523     & 0.483              \\ \hline
SmartEdit-7B             & 22.049   & 0.731 & 0.087 & 23.611     & \textbf{0.712}              & 25.258   & 0.742 & 0.055 & 20.950     & \textbf{0.789}              \\
SmartEdit-13B            & 23.596   & 0.751 & 0.068 & 23.536     & \textbf{0.771}              & 25.757   & 0.747 & 0.051 & 20.777     & \textbf{0.817}              \\ \hline
\end{tabular}
}
\caption{Quantitative comparison (PSNR$\uparrow$/SSIM$\uparrow$/LPIPS$\downarrow$/CLIP Score$\uparrow$ (ViT-L/14)/Ins-align$\uparrow$) on Reason-Edit. All the methods we compared have been fine-tuned using the same training data as that used by SmartEdit.}
% \vspace{-10pt}
\label{table:comparison}
\end{table*}

% Please add the following required packages to your document preamble:
% \usepackage{multirow}
\begin{table*}[]
\centering
\resizebox{1.0\textwidth}{!}{
\begin{tabular}{c|ccc|ccccc|ccccc}
\hline
\multirow{2}{*}{Exp ID} & \multirow{2}{*}{Plain} & \multirow{2}{*}{SimpleCA} & \multirow{2}{*}{BIM} & \multicolumn{5}{c|}{Understanding Scenarios} & \multicolumn{5}{c}{Reasoning Scenarios} \\ \cline{5-14} 
                        &                        &                           &                      & PSNR(dB)$\uparrow$ & SSIM$\uparrow$  & LPIPS$\downarrow$ & CLIP Score$\uparrow$ & \textbf{Ins-align$\uparrow$} & PSNR(dB) & SSIM  & LPIPS & CLIP Score & \textbf{Ins-align$\uparrow$} \\ \hline
1                       &           \ding{51}             &                           &                      & 20.975     & 0.713   & 0.108   & 23.36   & 0.695  & 23.848   & 0.725 & 0.074 & 20.33  & 0.694    \\
2                       &                        &         \ding{51}        &                      & 19.557     & 0.692   & 0.126   & 23.66    & 0.692   & 23.508   & 0.716 & 0.081 & 20.17 & 0.722      \\       
3                       &                        &                           &         \ding{51}         & 22.049     & 0.731   & 0.087   & 23.61  & \textbf{0.712}  & 25.258   & 0.742 & 0.055 & 20.95  & \textbf{0.789}    \\ \hline
\end{tabular}
}
\caption{Quantitative comparison (PSNR$\uparrow$/SSIM$\uparrow$/LPIPS$\downarrow$/CLIP Score$\uparrow$ (ViT-L/14)/Ins-align$\uparrow$) on Reason-Edit. These comparative experiments are conducted based on the SmartEdit-7B.
}
\label{table:bim}
\end{table*}

% Please add the following required packages to your document preamble:
% \usepackage{multirow}
\begin{table*}[]
\centering
\resizebox{1.0\textwidth}{!}{
\begin{tabular}{c|ccc|ccccc|ccccc}
\hline
\multirow{2}{*}{Exp ID} & \multirow{2}{*}{Edit} & \multirow{2}{*}{Segmentation} & \multirow{2}{*}{Synthetic editing dataset} & \multicolumn{5}{c|}{Understanding Scenarios} & \multicolumn{5}{c}{Reasoning Scenarios} \\ \cline{5-14} 
                        &                       &                               &                              & PSNR(dB)$\uparrow$ & SSIM$\uparrow$  & LPIPS$\downarrow$ & CLIP Score$\uparrow$ & \textbf{Ins-align$\uparrow$} & PSNR(dB) & SSIM  & LPIPS & CLIP Score & \textbf{Ins-align$\uparrow$} \\ \hline
1                       &           \ding{51}            &                               &                              & 17.568     & 0.664   & 0.171   & 22.79   & 0.201    & 22.400   & 0.706 & 0.102 & 19.22   & 0.233   \\
2                       &           \ding{51}            &         \ding{51}                      &                              & 18.960     & 0.690   & 0.143   & 22.83  &  0.361    & 21.774   & 0.693 & 0.116 & 19.82  & 0.311    \\
3                       &         \ding{51}              &                               &                  \ding{51}            & 19.562     & 0.702   & 0.111   & 22.32   & 0.440    & 23.595   & 0.715 & 0.079 & 20.43  & 0.567    \\
4                       &       \ding{51}                &          \ding{51}                     &                \ding{51}              & 22.049     & 0.731   & 0.087   & 23.61  &  \textbf{0.712}    & 25.258   & 0.742 & 0.055 & 20.95  & \textbf{0.789}    \\ \hline
\end{tabular}
}
\caption{Quantitative comparison (PSNR$\uparrow$/SSIM$\uparrow$/LPIPS$\downarrow$/CLIP Score$\uparrow$ (ViT-L/14)/Ins-align$\uparrow$) on Reason-Edit. These comparative experiments are conducted based on the SmartEdit-7B.
}
% \vspace{-10pt}
\label{table:dataset}
\end{table*}

\subsection{Comparison with State-of-the-Art Methods}

We compare SmartEdit with existing state-of-the-art instruction-based image editing methods, namely InstructPix2Pix, MagicBrush, and InstructDiffusion. Considering that these released models are trained on specific datasets, they would inevitably perform poorly if directly evaluated on Reason-Edit. To ensure a fair comparison, we fine-tune these methods on the same training set used by SmartEdit, and evaluate the fine-tuned models on Reason-Edit. The experimental results are shown in Tab~\ref{table:comparison}. From the quantitative results of the Reasoning Scenarios in the table, it can be observed that when we replace the clip text encoder in the diffusion model with LLaVA and adopt the proposed BIM module, both SmartEdit-7B and SmartEdit-13B achieve better results on these five metrics. This suggests that in scenarios requiring reasoning from instructions, a simple clip text encoder may struggle to understand the meaning of the instructions. However, the MLLM can fully utilize its powerful reasoning ability and world knowledge to correctly identify the corresponding objects and perform edits. 

The qualitative results further illustrate this point. As shown in Fig.~\ref{fig:comparison}, the first three examples are reasoning scenarios. In the first example, both SmartEdit-7B and SmartEdit-13B successfully identify the tool used for cutting fruit (knife) and remove it, while keeping the rest of the background unchanged. The second example can also be handled well by both of them. However, in the third example, we observe a difference in performance. Only SmartEdit-13B can accurately locate the object and perform the corresponding edits without altering other background areas. This suggests that in instruction-based image editing tasks that require reasoning, a more powerful MLLM model can effectively generalize its reasoning ability to this task. This observation aligns with the findings from LISA.

However, for understanding scenarios, we observe a difference in performance between SmartEdit-7B and SmartEdit-13B when compared to InstructDiffusion on the three metrics of PSNR/SSIM/LPIPS. Specifically, SmartEdit-7B performs worse than InstructDiffusion, while SmartEdit-13B outperforms InstructDiffusion on these metrics. Upon further analysis of the qualitative results, as shown in the $4_{th}$ and $5_{th}$ rows of Fig.~\ref{fig:comparison}, we find that from a visual perspective, both SmartEdit-7B and SmartEdit-13B appear superior to InstructDiffusion. This suggests that the three metrics do not always align with human visual perception. We confirm this phenomenon in the supplementary material (Section~\ref{sec:ins_align}). From the result of the Ins-align metric, it can be observed that SmartEdit shows a significant improvement compared to previous instruction-based image editing methods. Also, when adopting a more powerful MLLM model, SmartEdit-13B performs better than SmartEdit-7B on the Ins-align metric.

% To provide a more comprehensive evaluation of the effects of edited images, we propose a metric for assessing editing accuracy. This new metric aims to better evaluate the results produced by instruction editing methods. Detailed information about this metric is provided in the supplementary materials.

% In summary, SmartEdit demonstrates superior performance in quantitative metrics and qualitative results under reasoning scenarios, when compared to existing methods under fair training conditions. For understanding scenarios, our qualitative analysis reveals that images edited using SmartEdit align well with the instructions. This suggests that SmartEdit not only comprehends the instructions effectively but also can generate visually appealing results. This performance is superior when compared to existing instruction editing methods, indicating the effectiveness of SmartEdit in both reasoning and understanding scenarios.

\begin{figure}[t]
\centering
\small 
\begin{minipage}[t]{1.0\linewidth}
\centering
\includegraphics[width=1\columnwidth]{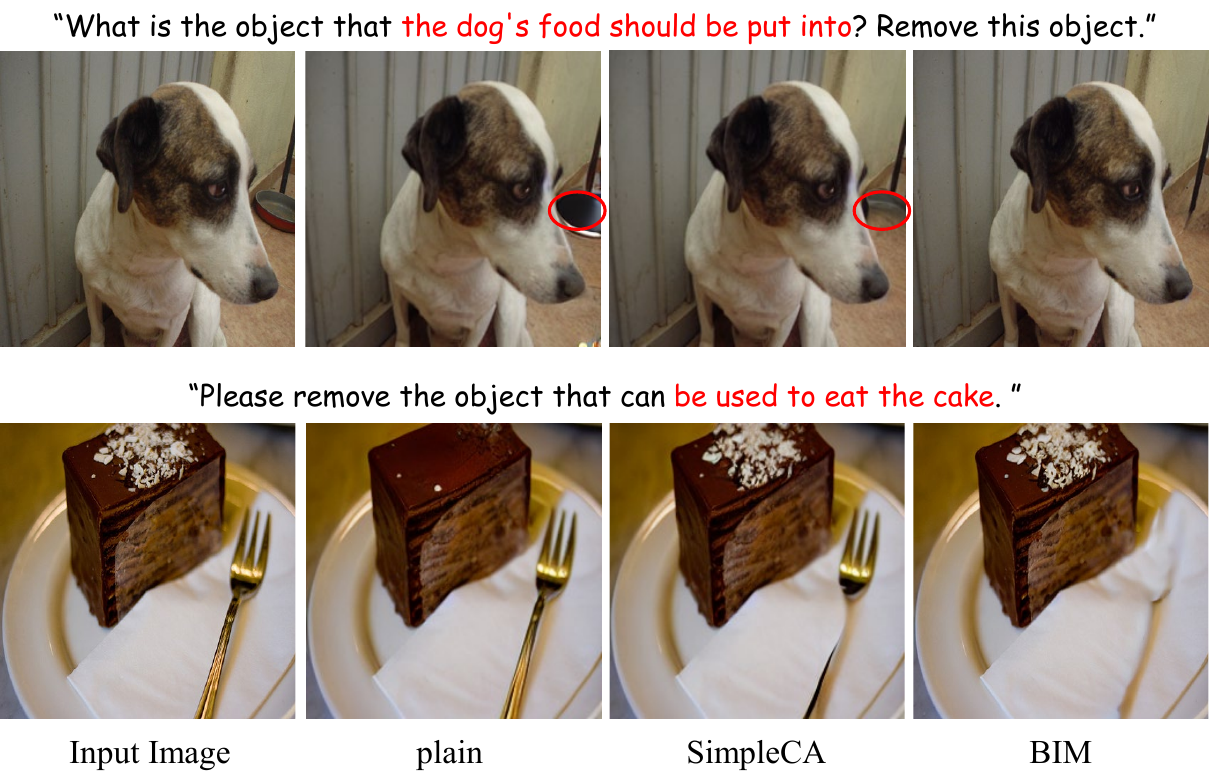}
\end{minipage}
\centering
% \vspace{-5pt}
\caption{The effectiveness of the BIM Module.}
% \vspace{0pt}
\label{fig:bim_effectiveness} 
\end{figure}

\begin{figure}[t]
\centering
\small 
\begin{minipage}[t]{1.0\linewidth}
\centering
\includegraphics[width=1\columnwidth]{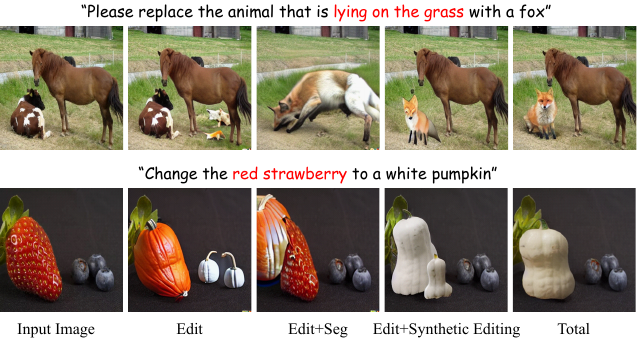}
\end{minipage}
\centering
% \vspace{-5pt}
\caption{The significance of joint training with multiple datasets.}
\vspace{-12pt}
\label{fig:dataset} 
\end{figure}

% \vspace{-2mm}
\subsection{Ablation Study on BIM}
% \vspace{-2mm}
\label{sec:ablation_bim}
To validate the effectiveness of the bidirectional information interaction in our proposed BIM module, we conduct comparative experiments on the SmartEdit-7B model. The details are presented in Tab.~\ref{table:bim}. The first experiment, denoted as Exp 1, aims to verify the necessity of the information interaction proposed in the BIM module. In this experiment, we remove the BIM module from the SmartEdit-7B model and directly apply the text feature output from QFormer to the diffusion model. The second experiment, denoted as Exp 2, aims to verify the necessity of the bidirectional information interaction proposed in the BIM module. Specifically, all blocks are discarded except for the cross-attention block on the image feature branch. Therefore, 
% replace the BIM module with simple cross attention (SimpleCA).  
the information from the text feature of QFormer is unidirectionally applied to the image feature. These two experiments are designed to test the impact of removing or altering the BIM module on the performance of SmartEdit-7B in complex understanding and reasoning scenarios. As shown in Tab.~\ref{table:bim}, if the BIM module is removed, there is a significant decline in all metrics for both understanding and reasoning scenarios. When the BIM module is replaced with the SimpleCA module, we observe a noticeable decline in all metrics, except for the clip score in understanding scenarios. Further comparison of the qualitative results in Fig.~\ref{fig:bim_effectiveness} confirms that the introduction of the BIM 
% and its bidirectional information interaction capability 
indeed enhances SmartEdit's instruction editing performance. To be specific, when we do not use the BIM module (i.e., plain), the dog bowl (first row) turns into other objects (marked with a red circle), and the fork (second row) does not change at all. After using SimpleCA, it can be found that the dog bowl and fork have been partially removed. When SmartEdit is equipped with BIM, the dog bowl and fork can be well removed.

\subsection{Ablation Study on Dataset Usage}
\label{sec:ablation_data}

In Section~\ref{method:traindata}, we explore an efficient strategy for data utilization, aiming to enhance SmartEdit's capabilities in handling complex understanding and reasoning scenarios. During the training process of SmartEdit, we employ the common editing dataset, segmentation dataset, and the synthetic editing dataset. To validate the significance of these different data types in boosting SmartEdit's performance, we conduct a series of ablation studies, as detailed in Tab.~\ref{table:dataset}. These experiments are based on the SmartEdit-7B model. In Exp 1, we train the model using only the editing data. In Exp 2, we incorporate segmentation data into the training process, building upon Exp 1. In Exp 3, we further add the synthetic editing data to the basis established in Exp 1. The quantitative results of these experiments reveal that segmentation data and synthetic editing data play complementary roles in enhancing the model's performance. This is further corroborated by the visual comparison in Fig.~\ref{fig:dataset}. For reasoning scenarios, when adopting only the editing dataset or combining the editing dataset and the segmentation dataset, the performance of SmartEdit is inferior. When the synthetic editing data is incorporated into the editing dataset, SmartEdit can accurately locate the specific objects. However, the output of SmartEdit is also mediocre (the generated fox has obvious artifacts, and two pumpkins are generated). When all these datasets are combined as the training set, the results generated by SmartEdit have a further significant improvement in visual effects. 

% for SmartEdit, the results generated by SmartEdit have a further significant improvement in visual effects.

% which clearly shows that the joint utilization of editing data, segmentation data, and synthetic editing data enables SmartEdit to deliver satisfactory results in complex understanding and reasoning scenarios.

\vspace{-2mm}
\section{Conclusion}
\label{sec:conclusion}
\vspace{-2mm}
In conclusion, this paper presents SmartEdit, a novel approach to instruction-based image editing that enhances understanding and reasoning capabilities by incorporating the Large Language Models (LLMs) with visual inputs. By introducing the Bidirectional Interaction Module (BIM), we have overcome challenges associated with the direct integration of LLMs and diffusion models in complex reasoning scenarios. Our data utilization strategy, which incorporates perception data and complex instruction editing data, effectively enhances SmartEdit's capabilities in handling complex understanding and reasoning scenarios. Evaluation on our newly constructed dataset, Reason-Edit, shows that SmartEdit outperforms previous methods, marking a significant step towards practical applications of complex instruction-based image editing.

% \section{Acknowledgment}

% \clearpage
{
    \small
    \bibliographystyle{ieeenat_fullname}
    \bibliography{main}
}

\clearpage
\clearpage
\setcounter{page}{1}
\maketitlesupplementary

% \section{Rationale}
% \label{sec:rationale}

In this supplementary file, we provide the following materials:

\begin{enumerate}
    \item Details of the data production pipeline.
    % \item Examples of the synthesized paired dataset (including images and corresponding instructions).
    % \item Examples of our collected Reason-Edit. (maybe not necessary)
    \item More quantitative comparisons on Reason-Edit.
    \item More visual results on Reason-Edit.
    \item Results of SmartEdit and other methods on MagicBrush.
    \item Difference between SmartEdit, MGIE~\cite{fu2023mgie} and InstructDiffusion~\cite{geng2023instructdiffusion}.
\end{enumerate}

\section{Details of the Data Production Pipeline}

As we mentioned in the main paper (Section 4.3), to effectively stimulate SmartEdit’s editing capabilities for more complex instructions, we synthesize approximately $476$ paired data as a supplement to the training data. This training dataset includes two major types of scenarios: complex understanding scenarios and reasoning scenarios. 

\begin{figure*}[t]
\centering
\small 
\begin{minipage}[t]{0.7\linewidth}
\centering
\includegraphics[width=1\columnwidth]{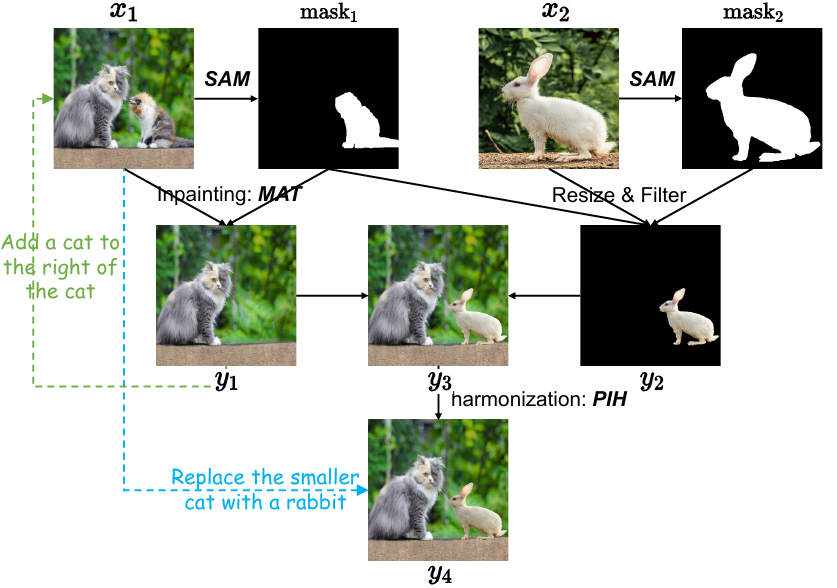}
\end{minipage}
\centering
% \vspace{2pt}
\caption{The data production pipeline of the synthetic paired training set (complex understanding scenarios). For $x_{1}$ and $x_{2}$, we first use SAM to generate $\mathrm{mask}_{1}$ and $\mathrm{mask}_{2}$. Then, we use MAT, combined with $x_{1}$ and $\mathrm{mask}_{1}$, to get $y_{1}$. At the same time, by performing specific operations on $\mathrm{mask}_{1}$, $\mathrm{mask}_{2}$, and $x_{2}$, we can get $y_{2}$. By combining $y_{1}$ and $y_{2}$, we can get $y_{3}$. Finally, we use the harmonization algorithm PIH to get $y_{4}$. ($y_{1}$, $x_{1}$, \textit{"Add a cat to the right of the cat"}) and ($x_{1}$, $y_{4}$, \textit{"Replace the smaller cat with a rabbit"}) can form the training samples.}
% \vspace{-10pt}
\label{fig:data_collection_pipe} 
\end{figure*}

\begin{figure*}[t]
\centering
\small 
\begin{minipage}[t]{0.9\linewidth}
\centering
\includegraphics[width=1\columnwidth]{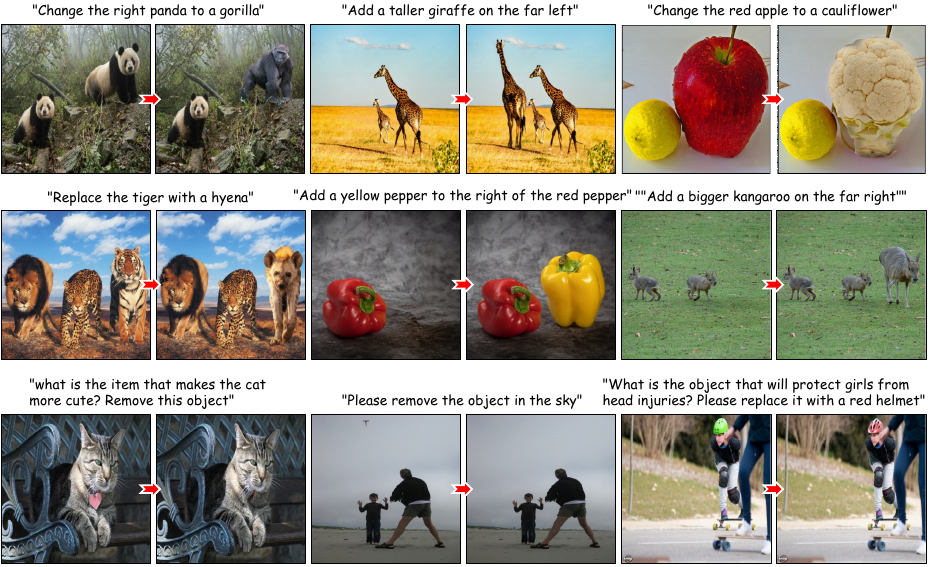}
\end{minipage}
\centering
\vspace{8pt}
\caption{Samples of complex understanding and reasoning scenarios in our synthesized paired training data. For each sample, the image on the left is the input image, and the image on the right is the image edited according to the instructions above.}
% \vspace{-10pt}
\label{fig:synthszied_paired_data}
\end{figure*}

For complex understanding scenarios, we establish a data production pipeline, which is illustrated in Fig.~\ref{fig:data_collection_pipe}. To be specific, We begin with two images, $x_{1}$ and $x_{2}$, collected from the internet. Using the SAM~\cite{kirillov2023segment} algorithm, we detect specific animals in these images. In image $x_{1}$, we identify a cat ($\mathrm{mask}_{1}$) that we aim to replace, and in $x_{2}$, we identify a rabbit ($\mathrm{mask}_{2}$) that we intend to use as a replacement. Following this, we apply the inpainting algorithm MAT~\cite{li2022mat} to $x_{1}$ and $\mathrm{mask}_{1}$, creating a new image, $y_{1}$, where the cat has been seamlessly removed. To prepare the rabbit from $x_{2}$ for insertion into $y_{1}$, we apply resize and filter operations to $\mathrm{mask}_{1}$, $\mathrm{mask}_{2}$, and $x_{2}$, resulting in a new image, $y_{2}$. We then merge $y_{1}$ and $y_{2}$ to form $y_{3}$, which features the rabbit in the place of the cat. Due to potential differences in saturation, contrast, and other parameters between $x_{1}$ and $x_{2}$, the rabbit may not blend well with the rest of the image. To rectify this, we apply the harmonization algorithm PIH~\cite{wang2023semi} to $y_{3}$ to obtain a more harmonious image, $y_{4}$. By utilizing some images in the entire process, we can obtain two pairs of training samples: where ($y_{1}$, $x_{1}$, \textit{"Add a cat to the right of the cat"}) can form one pair of training samples, with $y_{1}$ as the original image and $x_{1}$ as the ground truth; ($x_{1}$, $y_{4}$, \textit{"Replace the smaller cat with a rabbit"}) can also form a pair of training samples, with $x_{1}$ as the original image and $y_{4}$ as the ground truth. In Fig.~\ref{fig:synthszied_paired_data}, the first two rows show some complex understanding samples contained in the training data. 

For reasoning scenarios, we first generate the corresponding object's mask through SAM~\cite{kirillov2023segment}, then adopt stable diffusion~\cite{rombach2022high} to perform inpainting based on the provided instruction. Since the inpainting process can sometimes generate failure cases, we further manually filter the unsatisfied image. In the last row of Fig.~\ref{fig:synthszied_paired_data}, we illustrate some reasoning samples that are included in the training data.

\clearpage 

\clearpage
\section{More Quantitative Results on Reason-Edit}

\subsection{Instruction-Alignment Metric (Ins-align)} 
\label{sec:ins_align}
\begin{figure*}[t]
\centering
\small 
\begin{minipage}[t]{0.8\linewidth}
\centering
\includegraphics[width=1\columnwidth]{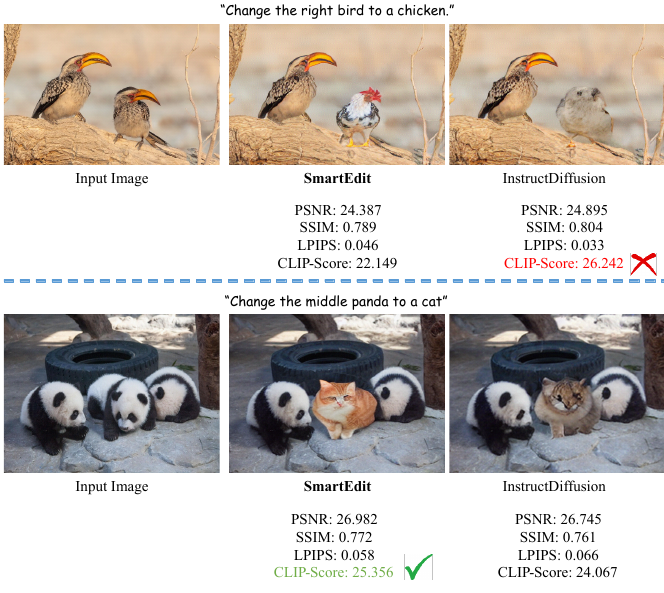}
\end{minipage}
\centering
\vspace{-10pt}
\caption{The evaluation of the outputs generated by SmartEdit and InstructDiffusion. }
\vspace{10pt}
\label{fig:smartedit_reasonedit_metric}
\end{figure*}

As mentioned in the main paper, PSNR/SSIM/LPIPS/CLIP-Score are the four most commonly used metrics in instruction-based image editing methods. For the foreground area, we calculate the CLIP Score~\cite{radford2021learning} between the foreground area of the edited image and the GT label. For the background area, we calculate the PSNR/SSIM/LPIPS~\cite{hore2010image, zhang2018unreasonable} between the edited image and the original input image. While these metrics can reflect the performance to a certain extent, they are not entirely accurate. This can be confirmed in Fig.~\ref{fig:smartedit_reasonedit_metric}. Specifically, in the first row of results, SmartEdit successfully generates a chicken, while InstructDiffusion does not generate a real chicken well. However, the CLIP-Score metric ranks InstructDiffusion higher. In the second row of images, the CLIP-Score aligns more with visual judgment, ranking SmartEdit's results higher. This indicates that the CLIP-Score metric may not always match human visual assessment. Regarding the PSNR/SSIM/LPIPS metrics, there is a significant variation in the results between SmartEdit and InstructDiffusion. Visually, the images edited by these two methods (the first row and the second row) do not have much visual difference in the background area, which indicates that these three metrics also cannot always accurately reflect the effectiveness of the instruction-based image editing methods. To provide a more accurate evaluation of the effects of edited images, we propose a metric for assessing editing accuracy. Specifically, we hire four workers to manually evaluate the results of these different methods on Reason-Edit. The evaluation criterion is whether the edited image aligns with the instruction. After obtaining the evaluation results from each worker, we average all the results to get the final metric result, which is Instruction-Alignment (Ins-align).

For all the experimental results in the main paper, we include the results of the Ins-align indicator, as shown in Tab.~\ref{table:comparison}, Tab.~\ref{table:bim}, and Tab.~\ref{table:dataset}. In Tab.~\ref{table:comparison}, we compare the results of SmartEdit with different existing instruction editing methods. It can be observed that when we use a metric consistent with human visual perception (Ins-align), for complex understanding and reasoning scenarios, SmartEdit shows a significant improvement compared to previous instruction-based image editing methods. Also, when adopting a more powerful LLM model, SmartEdit-13B performs better than SmartEdit-7B on the Ins-align metric.

Tab.~\ref{table:bim} and Tab.~\ref{table:dataset} present the results of the Ablation studies for BIM module and Dataset Usage, respectively. In Tab.~\ref{table:bim}, based on the results from the Ins-align metric, the introduction of the BIM module and its bidirectional information interaction capability indeed enhance SmartEdit’s instruction editing performance in complex understanding and reasoning scenarios. As shown in Tab.~\ref{table:dataset}, the joint utilization of editing data, segmentation data, and synthetic editing data enables SmartEdit to deliver better results in complex understanding and reasoning scenarios.

\subsection{User Study} 
\begin{figure*}[t]
\centering
\small 
\begin{minipage}[t]{0.8\linewidth}
\centering
\includegraphics[width=1\columnwidth]{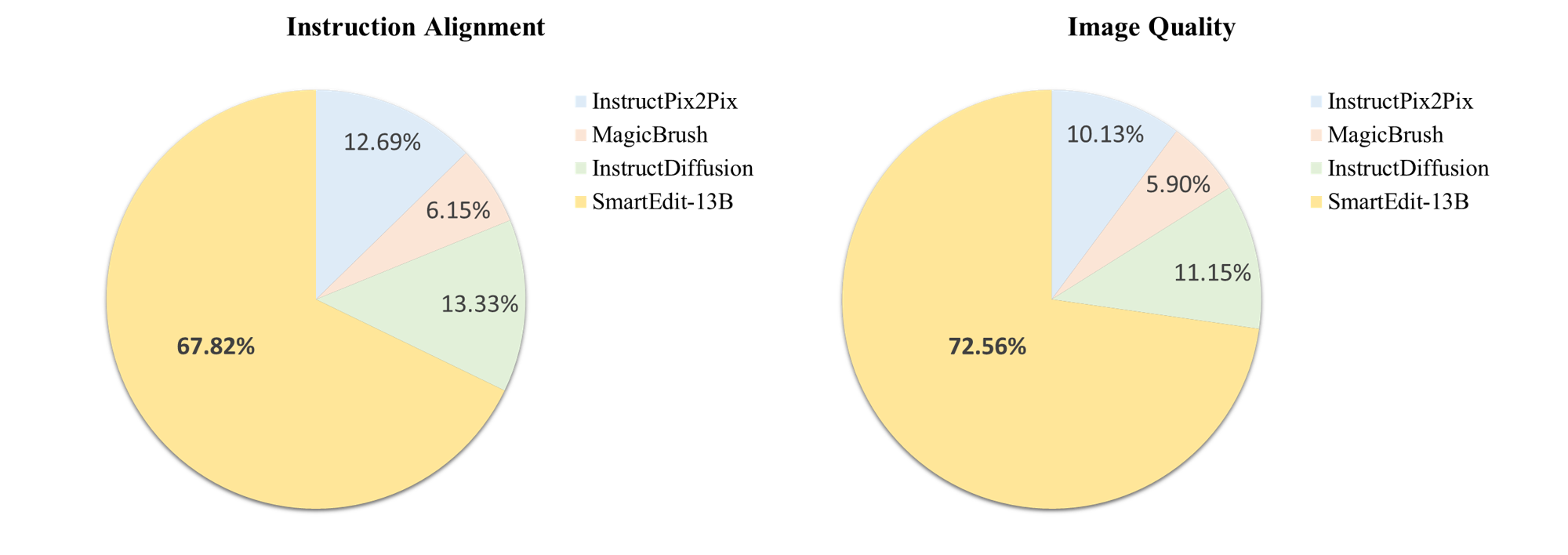}
\end{minipage}
\centering
% \vspace{2pt}
\caption{The results of user studies, comparing the results generated by InstructPix2Pix, MagicBrush, InstructDiffusion, and SmartEdit-13B. Based on the results from both the Instruction Alignment and Image Quality perspectives, SmartEdit demonstrates superior effectiveness.}
% \vspace{-10pt}
\label{fig:user_study}
\end{figure*}

To further verify the effectiveness of SmartEdit, we perform a user study. Specifically, we randomly select $30$ images from Reason-Edit, of which $15$ images belong to complex understanding scenarios, and the other $15$ belong to reasoning scenarios. For each image, we obtain the results of InstructPix2Pix, MagicBrush, InstructDiffusion, and SmartEdit, and randomly shuffle the order of these method results. As we mentioned in the main paper, for fairness, all comparison methods undergo fine-tuning on the same dataset as SmartEdit. In the end, we get 30 groups of images with shuffled order. For each set of images, we ask participants to independently select the two best pictures. The first one is the best picture corresponding to the instruction (i.e., Instruct-Alignment), and the second one is the picture with the highest visual quality under the condition of having editing effects (i.e., Image Quality). A total of $25$ people participate in the user study. The result is shown in Fig.~\ref{fig:user_study}. We can find that over $67\%$ of participants think that the effect of SmartEdit corresponds better with the instructions and more than $72\%$ of participants prefer the results generated by SmartEdit. This further suggests that SmartEdit is superior to other methods.

\clearpage

\clearpage
\section{More Visual Results on Reason-Edit}

\begin{figure*}[t]
\centering
\small 
\begin{minipage}[t]{1.0\linewidth}
\centering
\includegraphics[width=1\columnwidth]{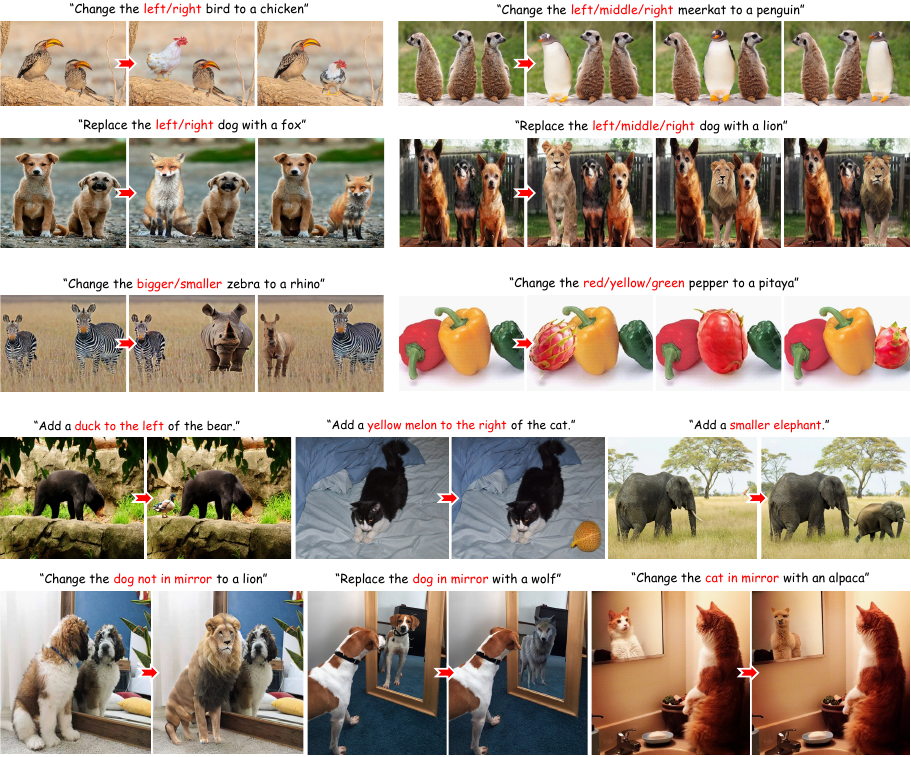}
\end{minipage}
\centering
% \vspace{2pt}
\caption{Visual effects of SmartEdit on Reason-Edit dataset (mainly on the complex understanding scenarios). It can be seen that for complex understanding scenarios (the instruction that contains various object attributes like location, relative size, color, and in or outside the mirror), SmartEdit has good instruction-based editing effects.}
% \vspace{-10pt}
\label{fig:smartedit_reasonedit}
\end{figure*}

\begin{figure*}[t]
\centering
\small 
\begin{minipage}[t]{0.9\linewidth}
\centering
\includegraphics[width=1\columnwidth]{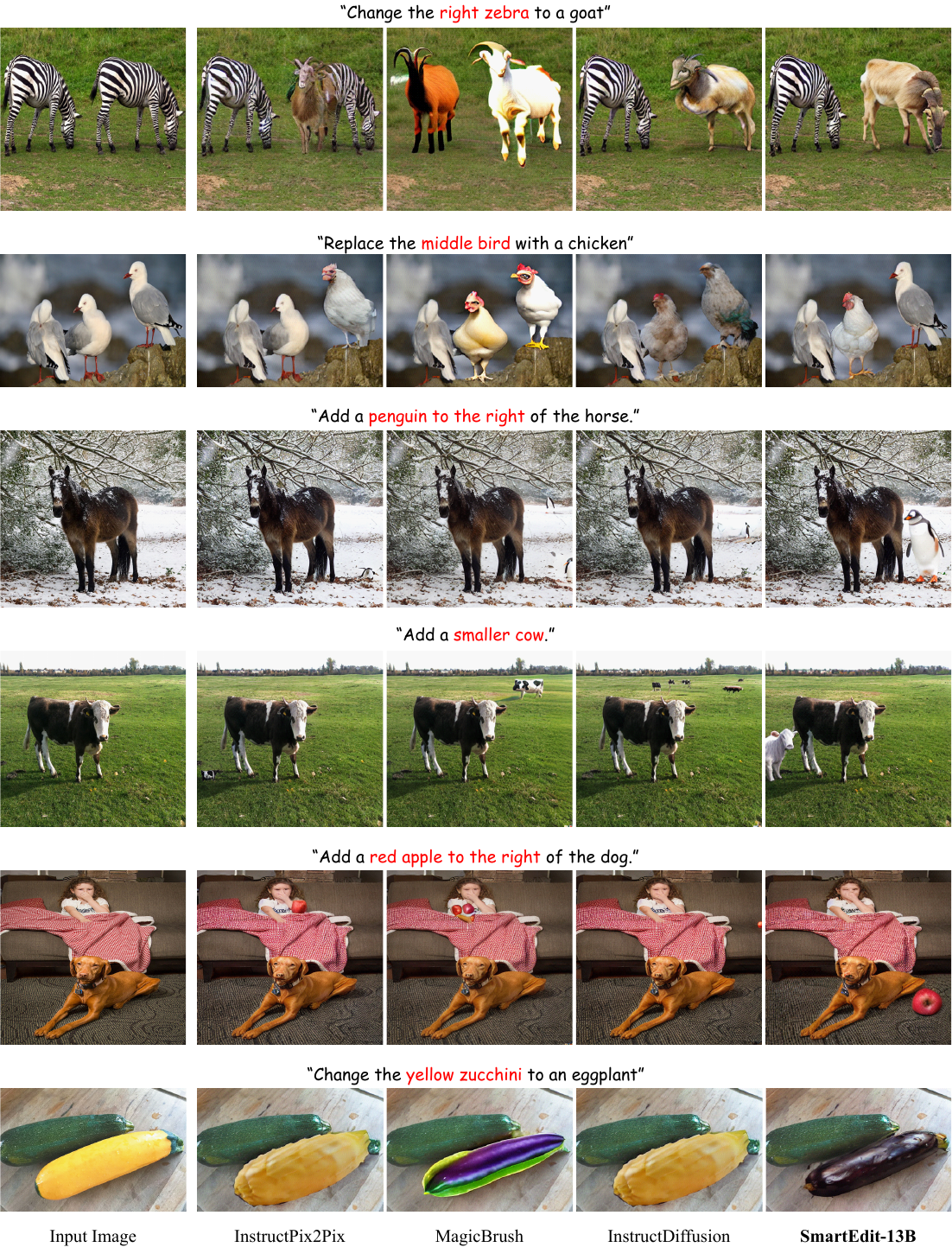}
\end{minipage}
\centering
% \vspace{2pt}
\caption{Qualitative comparison on Reason-Edit dataset (mainly on the complex understanding scenarios). Compared to other methods, SmartEdit can precisely edit specific objects in images according to instructions, while keeping the content in other areas unchanged.}
% \vspace{-10pt}
\label{fig:cmp_understanding}
\end{figure*}

For complex understanding scenarios, we show more editing results of SmartEdit in Fig.~\ref{fig:smartedit_reasonedit}. For the various object attributes, SmartEdit can understand the image and instructions well and can correctly edit the specified object accordingly. In addition, we compare the qualitative results of different methods for complex understanding scenarios, as shown in Fig.~\ref{fig:cmp_understanding}. From the first and second rows, it can be seen that InstructDiffusion can also edit specified objects according to instructions, but the quality of its edited images is much worse than that of SmartEdit. 
%Besides, for the second row, except SmartEdit, only SmartEdit correctly edited the bird in the middle. 
For the middle two rows of images, only MagicBrush among the existing methods understands the instructions and makes some modifications, but the image quality after editing is poor. For the last two rows of images, existing methods struggle to understand the instructions. SmartEdit, on the other hand, exhibits a superior ability to accomplish this task. 

\begin{figure*}[t]
\centering
\small 
\begin{minipage}[t]{0.95\linewidth}
\centering
\includegraphics[width=1\columnwidth]{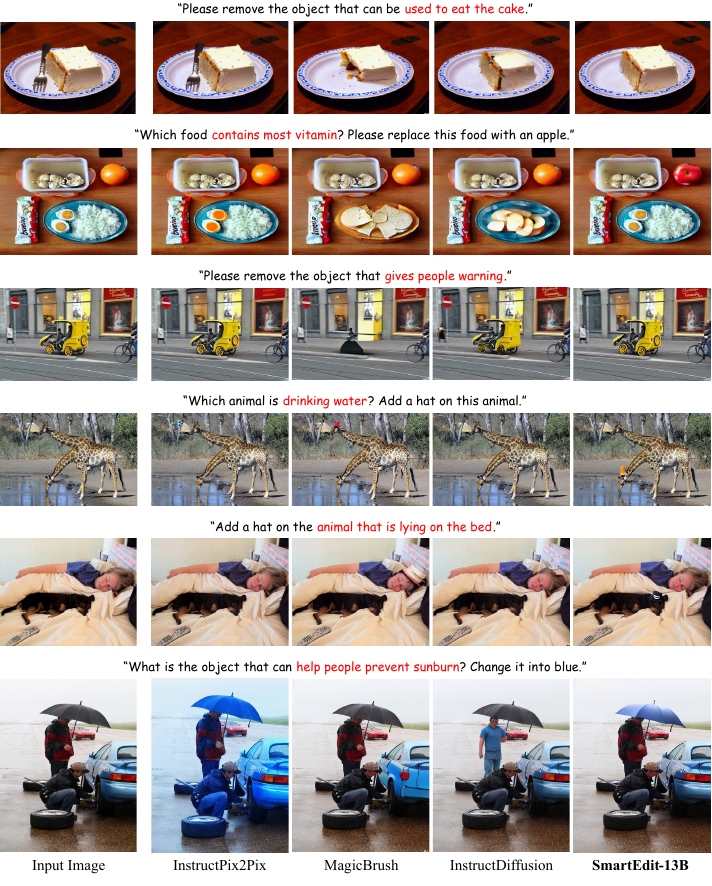}
\end{minipage}
\centering
% \vspace{2pt}
\caption{Qualitative comparison on Reason-Edit dataset (mainly on the complex reasoning scenarios). For reasoning scenarios, SmartEdit can effectively utilize the reasoning capabilities of the LLM to identify the corresponding objects, and then edit the objects according to the instructions. Other methods perform poorly in these scenarios.}
% \vspace{-10pt}
\label{fig:cmp_reasoning}
\end{figure*}

For reasoning scenarios, we provide a qualitative comparison of different methods on Reason-Edit, as shown in Fig.~\ref{fig:cmp_reasoning}. In the first row, although MagicBrush and InstructDiffusion can remove the fork, the part of the cake in the original image also gets modified accordingly. In contrast, SmartEdit not only removes the fork but also effectively protects other areas from being modified. For the second row, other methods do not find the food with the most vitamins (i.e., orange), but SmartEdit successfully identifies the orange and replaces it with an apple. From the third to the sixth rows, SmartEdit can understand the instructions and reason out the objects that need to be edited while keeping other areas unchanged. However, other methods struggle with understanding complex instructions and identifying the corresponding objects, leading to a poor editing effect. In summary, even though the existing methods use the same training data as SmartEdit for fine-tuning, the introduction of LLaVA and BIM modules enables the model to comprehend more complex instructions, thus yielding superior results.

\clearpage

\clearpage

\section{Results of SmartEdit and Other Methods on MagicBrush}

\begin{figure*}[t]
\centering
\small 
\begin{minipage}[t]{1.0\linewidth}
\centering
\includegraphics[width=1\columnwidth]{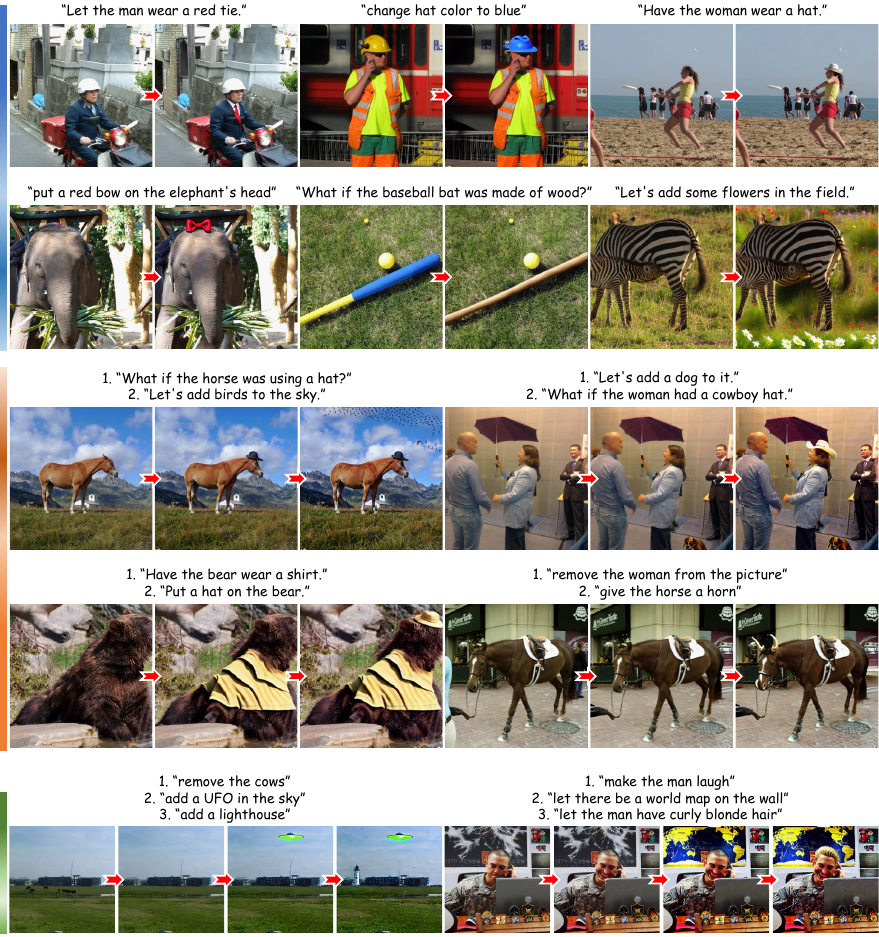}
\end{minipage}
\centering
% \vspace{2pt}
\caption{The performance of SmartEdit on the MagicBrush test dataset. SmartEdit has good editing effects on the MagicBrush test dataset, not only for single-turn but also for multi-turn.}
% \vspace{-10pt}
\label{fig:smartedit_magic}
\end{figure*}

In Fig.~\ref{fig:smartedit_magic}, we demonstrate the performance of SmartEdit on the MagicBrush~\cite{zhang2023magicbrush} test dataset. The first 2 rows are the editing results for single-turn, the middle 2 rows are for two-turn, and the last row is for three-turn. These results indicate that SmartEdit also has good editing effects on the MagicBrush test dataset, not only for single-turn, but also for multi-turn.

\begin{figure*}[t]
\centering
\small 
\begin{minipage}[t]{0.9\linewidth}
\centering
\includegraphics[width=1\columnwidth]{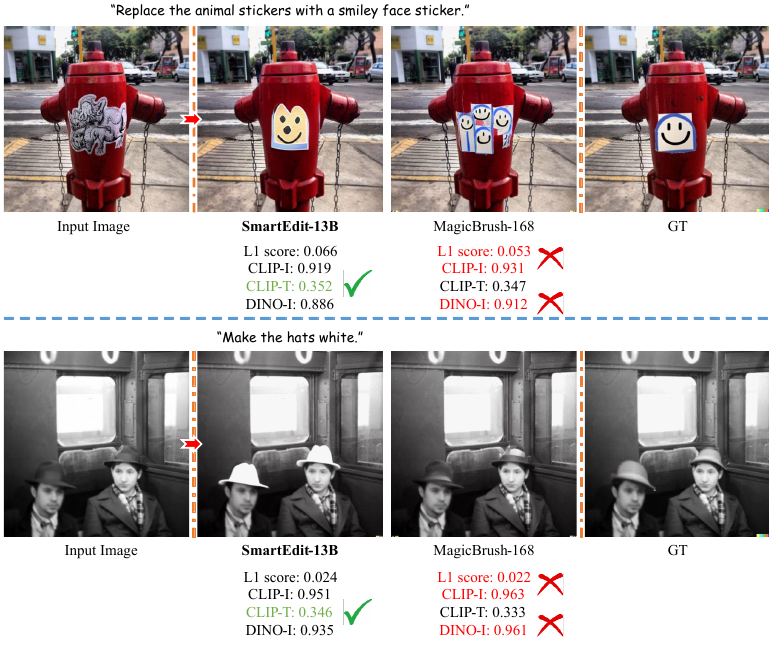}
\end{minipage}
\centering
% \vspace{2pt}
\caption{The evaluation of the outputs generated by SmartEdit and MagicBrush-168. Here we adopt these four metrics: $\mathrm{L}_{1}$, CLIP-I, CLIP-T, and DINO-I metrics. The results indicate that SmartEdit performs better than MagicBrush-168. However, it's important to note that the $\mathrm{L}_{1}$, CLIP-I, and DINO-I metrics may not correspond well with these results.}
% \vspace{-10pt}
\label{fig:magic_metric}
\end{figure*}

We further compare SmartEdit with other methods such as InstructPix2Pix~\cite{brooks2023instructpix2pix}, MagicBrush~\cite{zhang2023magicbrush}, and InstructDiffusion~\cite{geng2023instructdiffusion} on the MagicBrush test dataset. The quantitative results are presented in Tab.~\ref{table:compare_magic}. It's important to note that MagicBrush releases two distinct checkpoints, MagicBrush-52\footnote{\href{github repo}{https://huggingface.co/vinesmsuic/magicbrush-jul7}} (trained for $52$ epochs) and MagicBrush-168\footnote{\href{github repo}{https://huggingface.co/vinesmsuic/magicbrush-paper}} (trained for $168$ epochs). In the main paper of MagicBrush, the author utilizes MagicBrush-52 for qualitative results, while MagicBrush-168 is designed for quantitative results. As shown in Tab.~\ref{table:compare_magic}, MagicBrush-168 significantly outperforms MagicBrush-52 and other methods, including SmartEdit, in terms of metrics. However, upon further analysis of these metrics (as shown in Fig.~\ref{fig:magic_metric}), we find that the $\mathrm{L_{1}}$, CLIP-I, and DINO-I metrics may not be reliable. For instance, in the first set of images, SmartEdit effectively replaces the animal stickers with a smiley face sticker, while MagicBrush-168 adds multiple face stickers without completely removing the original animal stickers. Visually, SmartEdit's results appear superior to those of MagicBrush-168. A similar pattern is observed in the second set of images where SmartEdit successfully changes the hats of the two men in the original image to white, whereas MagicBrush-168 shows minimal changes. Despite this, the $\mathrm{L_{1}}$, CLIP-I, and DINO-I metrics indicate that MagicBrush-168's results are significantly better than SmartEdit's, suggesting that these metrics may not be a reliable measure of performance. In contrast, the CLIP-T metric seems to align more closely with the actual editing results, making it a potentially more reliable performance indicator. From Tab.~\ref{table:compare_magic}, it can be seen that SmartEdit performs better than MagicBrush-168 on the CLIP-T metric, while it is comparable to the results of MagicBrush-52.

\begin{figure*}[t]
\centering
\small 
\begin{minipage}[t]{0.9\linewidth}
\centering
\includegraphics[width=1\columnwidth]{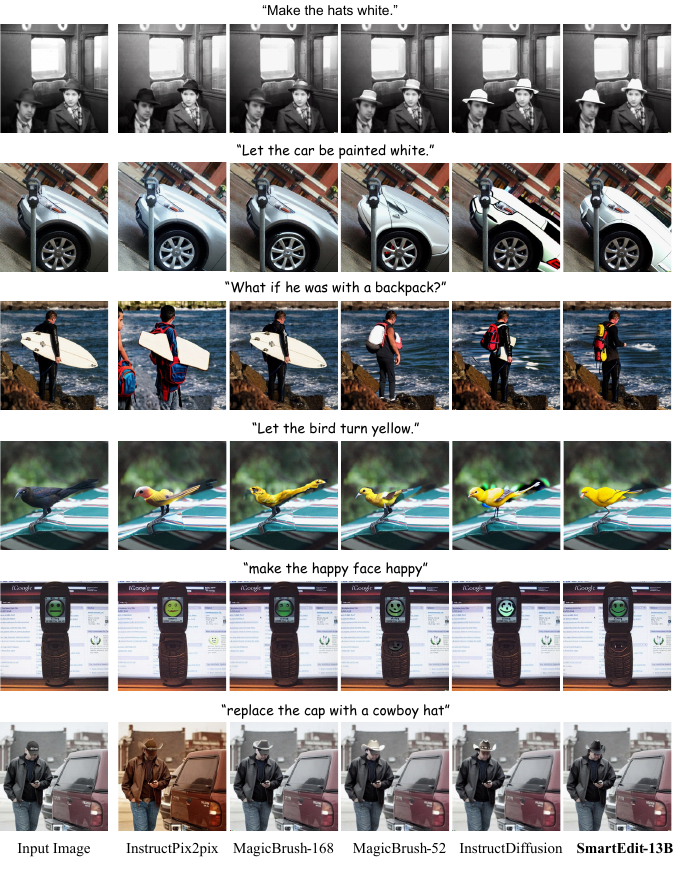}
\end{minipage}
\centering
% \vspace{2pt}
\caption{Qualitative comparison between our SmartEdit, MagicBrush-168, MagicBrush-52, InstructDiffusion, and InstructPix2Pix. Compared against other methods, SmartEdit effectively adheres to the instructions, showcasing superior results.}
% \vspace{-10pt}
\label{fig:compare_magic}
\end{figure*}

The comparative analysis of the qualitative results is illustrated in Fig.~\ref{fig:compare_magic}. InstructPix2Pix, which has not been trained on the MagicBrush dataset, demonstrates subpar performance. MagicBrush-168, in most cases, either tends to retain the original image (as seen in the first, second, third, and fifth rows) or exhibits poor editing results (as evident in the fourth and sixth rows). Although MagicBrush-52 shows better results than MagicBrush-168, the results after editing do not correspond well with the instructions (notably in the second and fourth rows). InstructDiffusion sometimes generates artifacts, as observed in the fourth and fifth rows. In contrast, SmartEdit effectively adheres to the instructions, showcasing superior results.

\begin{table}[h]
\centering
\resizebox{0.45\textwidth}{!}{
\begin{tabular}{c|cccc}
\hline
   Methods         & $\mathrm{L}_{1}$ $\downarrow$    & CLIP-I $\uparrow$ & CLIP-T $\uparrow$ & DINO-I $\uparrow$ \\ \hline
InstructPix2Pix     & 0.113 & 0.854  & 0.292  & 0.698 \\
MagicBrush-52     & 0.076 & 0.907  & 0.306  & 0.806  \\
MagicBrush-168    & 0.062 & 0.934  & 0.302  & 0.868  \\
InstructDiffusion & 0.097 & 0.892  & 0.302  & 0.777  \\
SmartEdit-7B      & 0.089 & 0.904  & 0.303  & 0.797  \\
SmartEdit-13B     & 0.081 & 0.914  & 0.305  & 0.815  \\ \hline
\end{tabular}
}
\caption{Quantitative comparison ($\mathrm{L}_{1}$/CLIP-I/CLIP-T/DINO-I) on the MagicBrush test set.}
\label{table:compare_magic}
\end{table}

\section{Difference between SmartEdit, MGIE and InstructDiffusion}

Recently, we have noticed a concurrent work: MGIE~\cite{fu2023mgie}. This method mainly uses MLLMs (i.e., LLaVA) to generate expressive instructions and provides explicit guidance for the following diffusion model. Compared with MGIE, there are three main differences. First, SmartEdit primarily targets complex understanding and reasoning scenarios, which are rarely mentioned in the MGIE paper. Secondly, in terms of network structure, we propose a Bidirectional Interaction Module (BIM) that enables comprehensive bidirectional information interactions between the image and the LLM output. Thirdly, we explore how to enhance the perception and reasoning capabilities of SmartEdit and propose a synthetic editing dataset. From both quantitative and qualitative results, it can be demonstrated that Our Smart has the ability to handle complex understanding and reasoning scenarios. 

Compared with InstructDiffusion, which proposes a unifying and generic framework for aligning computer vision tasks with human instructions, our primary focus is the field of instruction-based image editing. In our experiments, we find that the perceptual ability of the diffusion model is crucial for instruction editing methods. Since InstructDiffusion also trains on the segmentation dataset, for convenience, we directly use its weights as the initial weights for the diffusion model in SmartEdit. However, as can be seen from Fig.~\ref{fig:cmp_understanding} and Fig.~\ref{fig:cmp_reasoning}, despite InstructDiffusion utilizing a large amount of perception datasets for joint training, its performance in complex understanding and reasoning scenarios is somewhat standard. By integrating LLaVA and BIM module, and supplementing the training data with segmentation data and synthetic editing data, the final SmartEdit can achieve satisfactory results in complex understanding and reasoning scenarios.

\clearpage

% \clearpage

% WARNING: do not forget to delete the supplementary pages from your submission 
% \input{sec/X_suppl}

\end{document}